\documentclass[11pt]{article}

\PassOptionsToPackage{table}{xcolor}

\usepackage{preprint}

\usepackage{times}
\usepackage{latexsym}
\usepackage[T1]{fontenc}
\usepackage[utf8]{inputenc}
\usepackage{microtype}
\usepackage{inconsolata}

\usepackage{amsmath}
\usepackage{amssymb}
\usepackage{array}
\usepackage{arydshln}
\usepackage{booktabs}
\usepackage{enumitem}
\usepackage{fvextra}
\usepackage{graphicx}
\graphicspath{{./}}
\usepackage{placeins}
\usepackage{soul}
\usepackage{tabularx}
\usepackage[most]{tcolorbox}
\usepackage{stfloats}
\definecolor{lightgray}{RGB}{245, 245, 245}
\definecolor{wrongred}{RGB}{231, 76, 60}
\definecolor{headerblue}{HTML}{1F5AA6}
\definecolor{correctgreen}{HTML}{1B9E77}

\newcommand{\badge}[2]{%
  \begingroup
  \setlength{\fboxsep}{1.2pt}%
  \colorbox{#1!12}{\textcolor{#1}{\textbf{#2}}}%
  \endgroup
}
\newcommand{\HUMANbadge}{\badge{correctgreen}{HUMAN~\checkmark}}
\newcommand{\AIbadge}{\badge{wrongred}{AI~$\times$}}

\newcounter{prompttemplate}
\renewcommand{\theprompttemplate}{P\arabic{prompttemplate}}

\title{\textsc{Reader}: Reasoning-Enhanced AI-Generated Text Detection}

\author{
\begin{tabular*}{0.85\textwidth}{@{\extracolsep{\fill}}cccc@{}}
Pingfan Su$^{1,*}$ &
Kai Ye$^{1,*}$ &
Shijin Gong$^{2}$ &
Erhan Xu$^{1}$ 
\end{tabular*}
\\[0.6em]
\begin{tabular*}{0.75\textwidth}{@{\extracolsep{\fill}}ccc@{}}
Jin Zhu$^{3}$ &
Giulia Livieri$^{1,\dagger}$ &
Chengchun Shi$^{1,\dagger}$
\end{tabular*}
}

\begin{document}
\maketitle
\begingroup
\renewcommand{\thefootnote}{\fnsymbol{footnote}}
\footnotetext[1]{Equal contribution and listed in alphabetical order.}
\footnotetext[2]{Corresponding author.}
\endgroup

\begingroup
\renewcommand{\thefootnote}{\arabic{footnote}}
\footnotetext[1]{Department of Statistics, London School of Economics and Political Science; Emails: \texttt{p.su1@lse.ac.uk}, \texttt{K.Ye1@lse.ac.uk}, \texttt{E.Xu2@lse.ac.uk}, \texttt{G.Livieri@lse.ac.uk}, \texttt{c.shi7@lse.ac.uk}.}
\footnotetext[2]{School of Management, University of Science and Technology of China; Email: \texttt{shijin49@mail.ustc.edu.cn}.}
\footnotetext[3]{School of Mathematics, University of Birmingham; Email: \texttt{j.zhu.7@bham.ac.uk}.}
\endgroup

\begin{abstract}
Recent advances in large language models (LLMs) have made it increasingly difficult to distinguish human-written text from AI-generated content. Many existing detectors train supervised neural classifiers that achieve strong in-distribution performance but are often opaque and can degrade substantially under distribution shift.

We present \textsc{Reader}, a reasoning-enhanced AI text detector that outputs both a human/AI label and a structured rationale describing the evidence for its decision. A key component of our approach is \textsc{Read}, a curated supervision set of rationales and verdicts. We fine-tune an LLM on \textsc{Read} to build \textsc{Reader}, which reasons before detecting at inference time.

Despite having only \textbf{1.5B} parameters, \textsc{Reader} consistently outperforms existing detectors as well as prompted, high-capacity LLM baselines (GPT-5.2, Gemini-3-Pro, and DeepSeek-V3.2), which are \textbf{100} to \textbf{1000 times larger} in scale.

\end{abstract}
\begin{figure*}[t]
  \centering
  \includegraphics[width=0.95\textwidth]{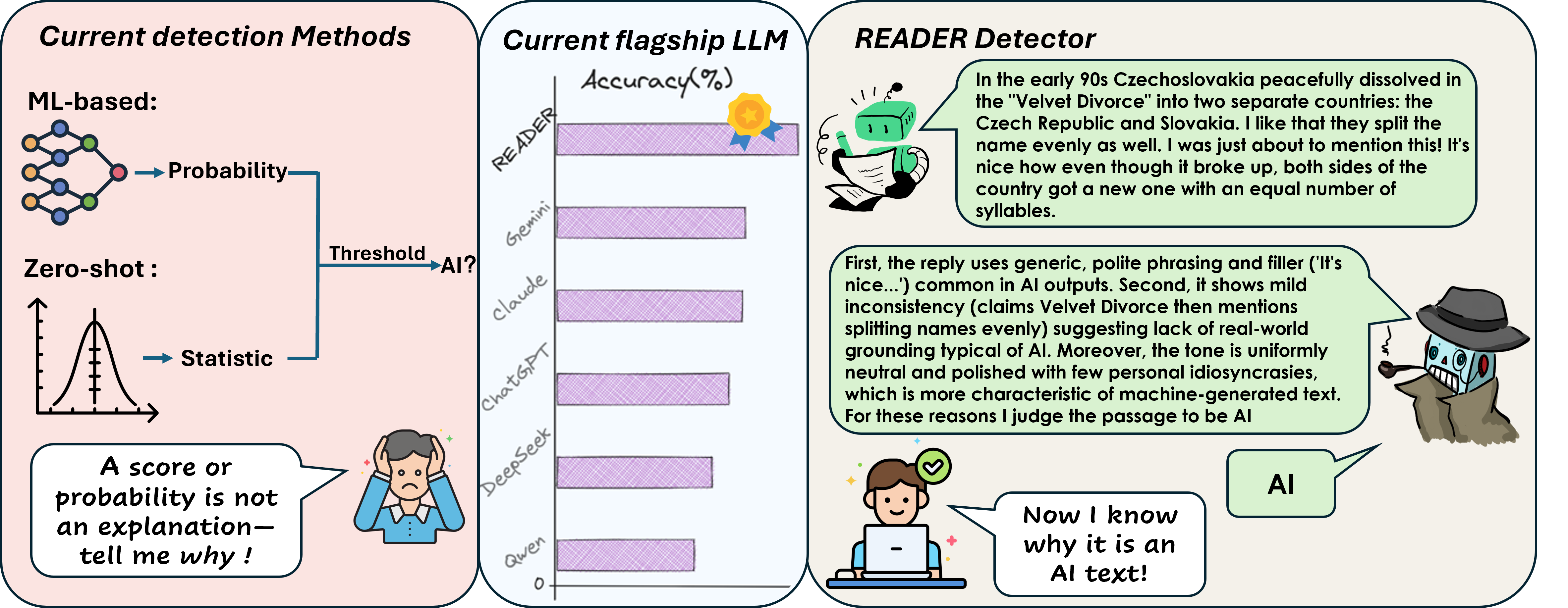}
  \caption{A comparison of \textsc{Reader} against existing detectors. \textbf{Left}: Conventional supervised and zero-shot detectors provide only a numerical score or probability, offering no interpretability. \textbf{Middle}: While high-capacity LLMs such as GPT and Gemini can be prompted to generate rationales, their detection accuracy remains substantially lower. \textbf{Right}: The proposed \textsc{Reader} provides both a high-accuracy classification label and a structured, human-readable reasoning trace.}
  \label{fig:compare}
\end{figure*}

\section{Introduction}
\label{sec:introduction}

Large language models (LLMs) increasingly produce fluent text that is often indistinguishable from human writing \citep[e.g.,][]{jakesch2023human, jones2025people}. While this capability enables many useful applications, it also raises practical concerns about misinformation, academic integrity, and trust in digital content  \citep[e.g.,][]{weidinger2021ethical, chen2024combating, cotton2024chatting}. As a result, detecting AI-generated text (AIGT) has become an important and fast-growing research area with broad real-world relevance.

Importantly, the need for AIGT detection extends beyond predictive accuracy. Under the EU AI Act, AI systems used in educational settings, including tools for ``monitoring and detecting prohibited behaviour of students during tests'', are categorized as \emph{high-risk} and are expected to satisfy transparency requirements \citep[e.g.,][]{eu_ai_act_annex3}. Similar expectations are emerging in scientific publishing, where stakeholders increasingly seek not only a binary outcome, but also a justification that can be interpreted and audited \citep[e.g.,][]{erol2025can}. In these settings, a detector's usability depends not just on performance, but also on whether its decisions can be understood and assessed.

Despite rapid progress, many AIGT detectors still operate as black boxes: they output a label (or score) without communicating the evidence behind it (see Section \ref{Positioning} and Appendix~\ref{Related_work} for reviews). Some work addresses this gap with post-hoc attribution tools such as LIME or SHAP \citep[e.g.,][]{mitrović2023chatgpthumandetectexplain, joshi2024hullmihumanvsllm, yuan2025emmm}.  
However, token-level attributions can be difficult for non-expert users to interpret, and prior work shows that post-hoc explanations can be manipulated to produce misleading rationales 
\citep[e.g.,][]{slack2020fooling}. 
These limitations motivate detectors that provide human-readable explanations together with their predictions, so that users can inspect the evidence stated by the detector rather than relying only on an opaque score.

Motivated by recent advances in LLM reasoning capabilities \citep[e.g.,][]{huang2023towards, plaat2024reasoning, liu2025fin, xu2025towards}, we explore a new paradigm: \emph{reasoning-enhanced detection}. Rather than attaching explanations after classification, we train a detector to produce an explicit, structured rationale before emitting its final verdict. The goal is to make the detector's stated evidence accessible in natural language, enabling decisions that can be inspected and reviewed.

We implement this idea with \textsc{Reader} (Figure~\ref{fig:compare}), a \textbf{R}easoning-\textbf{E}nhanced \textbf{A}I-text \textbf{DE}tecto\textbf{R} that outputs both (i) a human/AI label and (ii) a structured rationale describing the evidence for its decision. A key component is \textsc{Read}, a curated supervision set of \textbf{REA}soning \textbf{D}ata. We perform supervised fine-tuning (SFT) and group relative policy optimization \citep[GRPO,][]{shao2024deepseekmath} on \textsc{Read} so that \textsc{Reader} ``reasons before detecting" at inference time. Figure~\ref{fig:compare} summarizes this shift in detector design: moving from score-only or label-only outputs to decisions paired with structured, inspectable rationales.  

\begin{figure*}[t]
  \centering
  \includegraphics[width=0.95\textwidth]{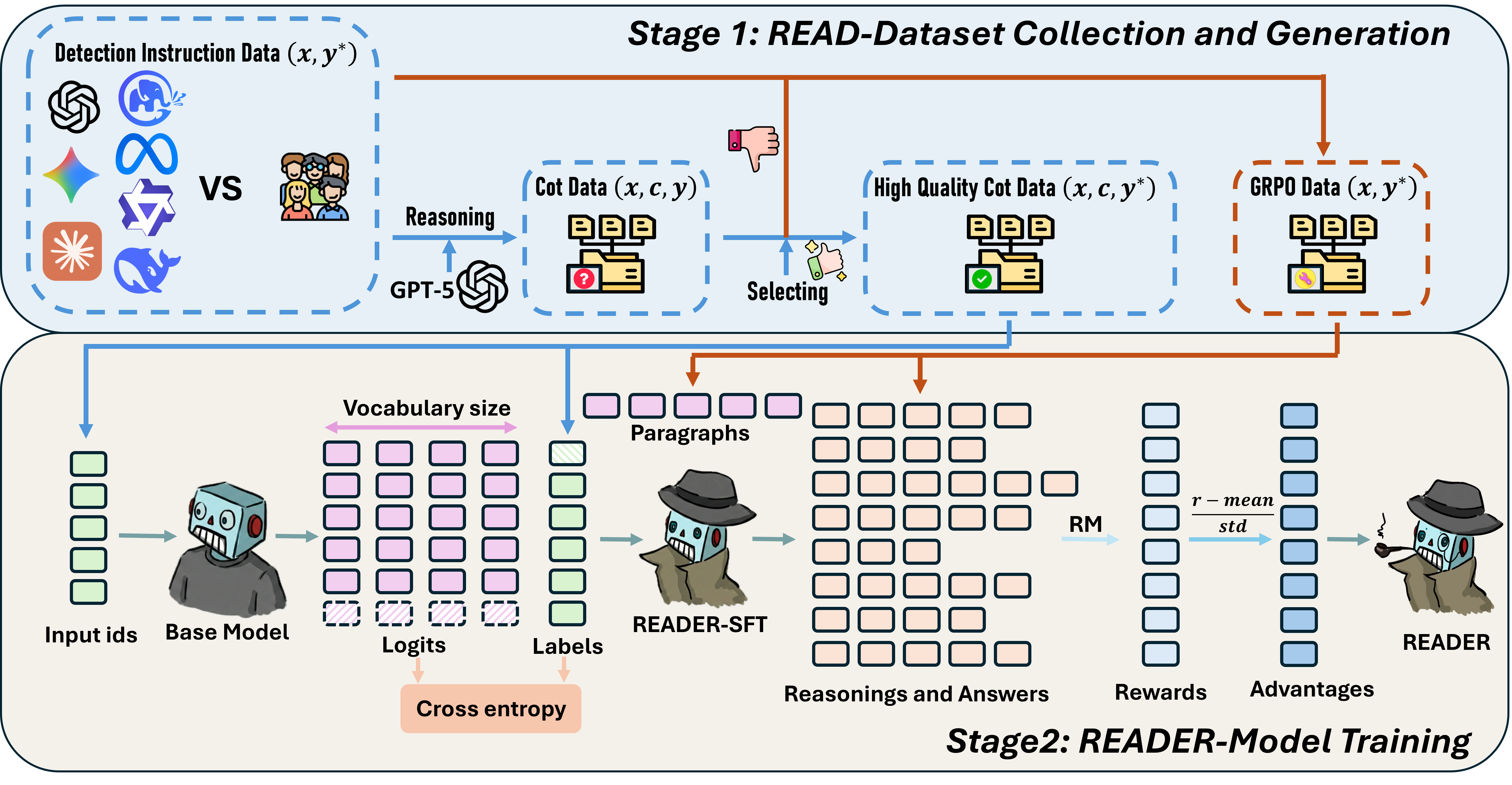}
  \caption{The \textsc{Reader} training pipeline. \textbf{Upper}: The first training stage constructs the \textsc{Read} dataset. We first collect a corpus of human-authored text and generate corresponding machine-text using various LLMs. Next, GPT-5 is utilized to generate reasoning traces that justify whether a given text is human- or LLM-authored. A rigorous filtering procedure is then applied to retain only high-quality instances for the second stage optimization. \textbf{Bottom}: The second training stage fine-tunes a base model on the \textsc{Read} dataset. More specifically, the model first undergoes SFT, followed by GRPO to build \textsc{Reader}.}
  \label{fig:pipeline}
\end{figure*}
This paper makes following three contributions: 
\begin{itemize}[leftmargin=*]

\item \textbf{The dataset \textsc{Read}.} We curate \textsc{Read}, a three-part resource for AIGT detection: reasoning-trace data tailored for explanatory supervision, answer-only data for standard detector training, and a challenging benchmark that pairs balanced human-written texts with AI-generated counterparts from multiple state-of-the-art LLMs across diverse domains.

\item \textbf{The detector \textsc{Reader}.} Using Qwen2.5-1.5B-Instruct as the base model, we fine-tune on \textsc{Read} to obtain \textsc{Reader} that produces transparent, step-by-step justifications alongside its human/AI verdict, without requiring additional fine-tuning when transferring to out-of-domain or out-of-distribution (OOD) texts.

\item \textbf{Strong empirical performance.} Across both in-distribution and OOD settings, \textsc{Reader} improves detection accuracy over strong supervised detectors and prompted general-purpose LLM baselines. In particular:
\begin{enumerate}[leftmargin=*]
    \item Compared to most existing zero-shot or supervised detectors, \textsc{Reader} achieves an absolute improvement of over \textbf{10-30\%} in detection accuracy (Table~\ref{tab:oracle_threshold_baselines_best2}), generalizes exceptionally well to out-of-domain or out-of-distribution data (Table \ref{tab:blackbox_realworld}), and handles short passages more effectively (Figure \ref{fig:prompt_len_acc}).
    \item Compared with state-of-the-art general-purpose LLMs (including GPT-5.2, Claude-Sonnet-4.5, DeepSeek-V3.2, and Gemini-3-Pro), \textsc{Reader} achieves an absolute improvement of over \textbf{20\%}  points in accuracy (Table~\ref{tab:accuracy_format_unusable_rates}) while providing transparent rationales that support its classifications (Table~\ref{tab:comparison-example}), despite being \textbf{100} to \textbf{1,000} times smaller than these models.
    \item \textsc{Reader} remains robust to multiple types of adversarial attacks (Appendix~\ref{app:ood-details}), and its generated rationales generally support the final verdict (Appendix~\ref{app:rationale_coupling}). 
   
\end{enumerate}
\end{itemize}
\subsection{Positioning within AIGT Detection}
\label{Positioning}
Existing AIGT detectors can be broadly categorized into three groups: (i) supervised detectors, (ii) zero-shot detectors, and (iii) watermarking-based detectors \citep[e.g.,][]{solaiman2019release, mitchell2023detectgpt, kirchenbauer2023watermark}. Supervised and zero-shot approaches are passive: they infer authorship from the observed text using learned classifiers, likelihood-based statistics, perturbation signals, rewrite consistency, or other text-derived features \citep[e.g.,][]{gehrmann2019gltr, bao2024fastdetectgpt, guo2024biscope, hao2025learning}. Watermarking methods, in contrast, require control over the generation process~\citep[e.g.,][]{kirchenbauer2023watermark, dathathri2024scalable,  huo2024token}. \textsc{Reader} is designed for the passive setting, where only the final text is available. 

Within passive detection, most existing systems return only a score, probability, or binary label, providing limited support for users who need to inspect the evidence behind a decision. Explainability-oriented detectors partially address this limitation, but typically rely on post-hoc attribution or visualization methods, which can be difficult to interpret and may not align with the detector's decision process~\citep[e.g.,][]{mitrovic2023chatgpt, gehrmann2019gltr, ji2025detectingmachinegeneratedtextsjust}. \textsc{Reader} instead trains a compact LLM to produce a structured rationale before its final verdict. 

In summary, \textsc{Reader} is positioned as a reasoning-enhanced passive detector: like conventional AIGT detectors, it requires only the input text for detection, but it additionally provides explicit natural-language rationales that make its decisions more explainable. A more detailed review of the literature is provided in Appendix~\ref{Related_work}.

\section{Methodology}
This section is organized as follows. We first detail the construction of the \textsc{Read} dataset (Section \ref{READ}). We next present the training of our \textsc{Reader} detector (Section \ref{sec:READERtraining}). Our complete pipeline is visualized in Figure \ref{fig:pipeline}.

\subsection{\textsc{Read} Construction} \label{READ}
The \textsc{Read} construction proceeds in four steps:
(i) collecting detection instruction data and reserving a held-out test split immediately. (ii) generating reasoning traces via a teacher model; (iii) pruning the dataset to retain high-quality instances; (iv) splitting into SFT and GRPO training data subsets. Figure~\ref{fig:pipeline} (upper panel) illustrates the construction of the \textsc{Read} training set. 

\smallskip

\noindent \textbf{Step 1: Detection instruction data curation.}
We construct an instruction dataset $\mathcal{D}_{\text{inst}}=\{(x_i, y_i^{*})\}_{i=1}^n$, where each text $x_i$ is paired with a ground-truth label $y_i^{*} \in \{\texttt{Human}, \texttt{AI}\}$ representing its underlying authorship. 

The human-authored texts are collected from two public datasets. The first is \textit{MAGE} \citep{li2024mage}, which contains data across 10 domains: (i) \ul{\textit{ChangeMyView (CMV)}} for Reddit discussions \citep{tan2016winning}; (ii) \ul{\textit{ELI5}} for open-domain question answering \citep{fan2019eli5longformquestion}; (iii)  \ul{\textit{HellaSwag}} for commonsense reasoning \citep{zellers2019hellaswagmachinereallyfinish}; (iv)  \ul{\textit{ROCStories Corpora}} (ROC) for story completion \citep{mostafazadeh2016corpusevaluationframeworkdeeper}; (v) \ul{\textit{SciXGen}} (SCI) for scientific writing \citep{chen2021scixgenscientificpaperdataset}; (vi)  \ul{\textit{SQuAD}} for Wikipedia-style question answering \citep{rajpurkar2016squad}; (vii) \ul{\textit{TLDR}}\footnote{\url{https://huggingface.co/datasets/JulesBelveze/tldr_news}} for news articles; (viii) \ul{\textit{WritingPrompts}} (WP) for open-ended writing generation~\cite{fan2018hierarchical}; (ix) \ul{\textit{XSum}} for news summarization \citep{narayan2018don}; and (x) \ul{\textit{Yelp}} for business reviews \citep{zhang2015character}. The second dataset is \textit{Rewrite} \citep{hao2025learning}, which paraphrases human texts across 21 sub-domains. 

The two datasets also provide AI-generated texts from the following five LLMs: \ul{\textit{GLM-130B}}, \ul{\textit{GPT-3.5-Turbo}}, \ul{\textit{LLaMA-65B}}, \ul{\textit{OPT-30B}}, and \ul{\textit{Gemini-1.5-Pro}}. To improve model coverage, we adopt the prompting protocol in \citet{hao2025learning} and collect additional data from three more recent models (\ul{\textit{Claude-Haiku-4.5}}, \ul{\textit{Qwen3-Flash}}, and \ul{\textit{DeepSeek-V3.2}}).

\smallskip

\noindent \textbf{Train/test split}. After constructing $\mathcal{D}_{\text{inst}}$, we split it into a training pool $\mathcal{D}_{\text{train}}$ and a held-out test set $\mathcal{D}_{\text{test}}$. The test set is balanced between human- and AI-authored texts and is used only for evaluation. All subsequent rationale generation, filtering, and training data construction are performed only on $\mathcal{D}_{\text{train}}$.

\smallskip

\noindent \textbf{Step 2: Teacher prompting (rationale generation).}
We sample a subset $\mathcal{D}_{\text{sub}}$ from $\mathcal{D}_{\text{train}}$ and use GPT-5 as a high-capacity teacher to generate, for each $x_i \in \mathcal{D}_{\text{sub}}$, a predicted authorship label $\widehat{y}_i$ and a short structured rationale $c_i$ explaining the prediction. The output is required to follow a standardized format, consisting of a few rationale sentences followed by a final label token. To ensure the quality of the teaching rationales, we select the final teacher prompt through human filtering and independent LLM-as-a-judge protocol evaluation~\cite{gu2024survey}. Details of the prompt-selection procedure and the full prompts are provided in Appendix~\ref{prompts}.

\smallskip

\noindent \textbf{Step 3: Filtering.}
After controlling rationale quality at the prompt level, we further apply stricter instance-level filtering to construct the final supervised training set. Specifically, we retain only examples that satisfy two objective criteria: (i) the teacher output strictly follows the specified format, containing a reasoning part followed by a classification label; and (ii) the predicted label $\widehat{y}_i$ matches the ground-truth label $y_i^*$. This filtering step removes malformed outputs, parsing failures, and teacher predictions that are inconsistent with the dataset annotation. See some representative retained and filtered examples in Appendix~\ref{app:filtering}, Figures~\ref{fig:filter-correct-human}--\ref{fig:filter-extra-closing}.

\smallskip

\noindent \textbf{Step 4: Data splitting.} The filtered high-quality samples form our SFT dataset $\mathcal{D}_{\text{SFT}}=\{(x_i, c_i, y_i^{*})\}_{i=1}^{N}$. The remaining samples where the teacher model failed to predict the label correctly or provided generic rationales are collected as a set of ``challenging'' instances. By stripping these failed rationales and merging them with the remaining text-label pairs from the original pool (i.e., $\mathcal{D}_{\text{train}} \setminus \mathcal{D}_{\text{sub}}$), we construct the GRPO dataset, $\mathcal{D}_{\text{GRPO}}=\{(x_i, y_i^{*})\}_{i=1}^{M}$. Both datasets are utilized in the subsequent two-step training: the model is first guided to learn how to reason via SFT, and next optimized via GRPO to enhance its detection capacity.

Finally, the training set consists of \textbf{19,684} high-quality rationale data and \textbf{77,103 } answer-only instances. The test set contains \textbf{9,600} samples. A detailed breakdown of samples across domains is reported in Table~\ref{tab:data_distribution}.

\begin{table}[t]
\centering
\small
\setlength{\tabcolsep}{2.pt}
\caption{Sample sizes by domain for the \textsc{Read} dataset.}
\label{tab:data_distribution}
\rowcolors{2}{blue!3}{white}
  \begin{center}
    \begin{small}
      \begin{sc}
\begin{tabular}{@{}l | c c c | c@{}}
\toprule
\rowcolor{white}
& \multicolumn{4}{c}{\textsc{Read}} \\
\rowcolor{white}
\textbf{Domain} & \textbf{SFT} & \textbf{GRPO} & \textbf{Test} & \textbf{Total} \\
\midrule
CMV & 1,765 & 5,975 & 641 & 8,381 \\
ELI5 & 1,725 & 8,469 & 779 & 10,973 \\
HellaSwag & 1,552 & 6,297 & 778 & 8,627 \\
ROC & 1,380 & 6,689 & 805 & 8,874 \\
SCI & 1,763 & 4,283 & 509 & 6,555 \\
SQuAD & 1,692 & 5,040 & 551 & 7,283 \\
TLDR & 1,393 & 5,147 & 630 & 7,170 \\
WP & 1,766 & 7,797 & 798 & 10,361 \\
XSum & 1,721 & 7,801 & 771 & 10,293 \\
Yelp & 1,485 & 5,735 & 560 & 7,780 \\
Rewrite & 3,442 & 13,870 & 2,778 & 20,090 \\
\midrule
\rowcolor{white}
\textbf{Total} & \textbf{19,684} & \textbf{77,103} & \textbf{9,600} & \textbf{106,387} \\
\rowcolor{white}
\textbf{Human/AI} & 50.4/49.6 & 42.3/57.7 & 50.0/50.0 & 44.5/55.5 \\
\bottomrule
\end{tabular}
      \end{sc}
    \end{small}
  \end{center}
\end{table}

\subsection{\textsc{Reader} Training}\label{sec:READERtraining}
Figure~\ref{fig:pipeline} (bottom panel) depicts our two-step training paradigm for \textsc{Reader}.

\smallskip

\noindent \textbf{Step 1: SFT}. We perform supervised fine-tuning on a 1.5B-parameter Qwen-Instruct base model using the high-quality demonstrations in $\mathcal{D}_{\text{SFT}}$ to instill reasoning capabilities. Specifically, we minimize the  cross-entropy loss for fine-tuning the model parameter $\theta$: 
\[
\begin{aligned}
\mathcal{L}_{\text{SFT}}(\theta)
= -\mathbb{E}_{(x,c,y^{*})\sim \mathcal{D}_{\text{SFT}}}
\bigl[
&\log p_{\theta}(c\mid x) \\
&+ \log p_{\theta}(y^* \mid x, c)
\bigr].
\end{aligned}
\]
where $p_{\theta}$ represents the conditional probability of a completion given its prefix. By training on the high-quality demonstrations in $\mathcal{D}_{\text{SFT}}$, the resulting model, \textsc{Reader}-SFT, learns to generate coherent detection rationales followed by a final classification decision. 

\smallskip

\noindent \textbf{Step 2: GRPO}. To further enhance the model's reasoning capabilities, we apply GRPO, a ``critic-free" variant of the proximal policy optimization algorithm \citep{schulman2017proximal},  to fine-tune \textsc{Reader}-SFT on the GRPO dataset $\mathcal{D}_{\text{GRPO}}=\{(x_i, y_i^{*})\}_{i=1}^{M}$. Specifically, for each $x_i$, we sample multiple reasoning traces and compute their relative rewards within the group to approximate the advantage function. This bypasses the need for learning a complex, neural-network-parameterized critic and substantially facilitates the computation. We next utilize these advantage estimates to construct the policy value estimator and update $\theta$ via policy gradient. 
Our ablation study demonstrates that GRPO substantially improves the reasoning capacity of \textsc{Reader} compared to \textsc{Reader}-SFT (Appendix~\ref{sec:ablation}).

\section{Experiments}
\label{sec:Experiments}
In this section, we conduct extensive experiments to evaluate the effectiveness and robustness of \textsc{Reader} against \textbf{15} existing detectors and \textbf{5} high-capacity general-purpose LLMs. 

\smallskip

\noindent \textbf{Summary}. We begin with a high-level overview of our empirical findings: \vspace{-0.5em}
\begin{itemize}[leftmargin=*]
    \item \textit{\textbf{In-distribution evaluation} (Section \ref{subsec:indis} and Appendix~\ref{app:extended-read-results})}: We compare \textsc{Reader} against state-of-the-art zero-shot and supervised detectors using the \textsc{Read} test set. This comparison demonstrates that (i) \textsc{Reader} is highly effective, achieving an absolute reduction in detection error of \textbf{10--30\%} over most existing baselines. (ii) Unlike many detectors that are sensitive to the choice of classification thresholds or passage length, \textsc{Reader} is threshold-agnostic while remaining robust to text length. \vspace{-0.5em}
    \item \textit{\textbf{Out-of-distribution evaluation} (Section \ref{subsec:outofdis} and Appendix~\ref{app:ood-details})}: We evaluate \textsc{Reader} on out-of-distribution data to assess its generalizability. The results show that \ul{\textit{\textsc{Reader} generalizes exceptionally well and remains robust to multiple types of adversarial attacks}}, often maintaining its in-distribution performance. In contrast, the performance of many supervised detectors degrades sharply when encountering distribution shifts.\vspace{-0.5em}
    \item \textit{\textbf{Benchmarking against general-purpose LLMs} (Section \ref{subsec:generalLLM})}: We compare \textsc{Reader} against state-of-the-art general-purpose LLMs including GPT-5.2, Gemini-3-Pro, DeepSeek-V3.2, and find that \textsc{Reader} achieves an absolute improvement of \textbf{20}--\textbf{40\%} in detection accuracy, while being \textbf{100} to \textbf{1,000} smaller in size.\vspace{-0.5em} 
    \item \textit{\textbf{Ablation} (Appendix~\ref{sec:ablation})}: We compare the base model, \textsc{Reader-SFT}, and \textsc{Reader}, and find that SFT and GRPO substantially improve detection performance. We also observe that chain-of-thought (CoT) inference improves accuracy for both \textsc{Reader-SFT} and \textsc{Reader}. \vspace{-0.5em}
    \item \textit{\textbf{Rationale validation}} (Appendix~\ref{app:rationale_coupling}): Our analysis further shows that \textsc{Reader}'s generated rationales contain interpretable evidence and are strongly aligned with its final verdicts.\vspace{-0.5em} 
\end{itemize}
These results support the effectiveness of the proposed reasoning-enhanced detection pipeline. 
Before detailing the experimental results, we describe our evaluation datasets and specify the baseline detectors.

\smallskip

\noindent \textbf{Dataset}. We utilize the test split of \textsc{Read}, as described in Section~\ref{READ}, for our in-distribution evaluation. The test set is highly diverse, covering all domains. Furthermore, it is balanced to include approximately equal numbers of AI-generated texts from different models, with the total amount of AIGT being similar to that of human-authored text. 

We construct three complementary OOD test sets to evaluate the robustness of \textsc{Reader} beyond the training distribution: a generator-level OOD set following \citet{bao2024fastdetectgpt}, a cross-family and cross-domain stress test set, and a third test set for evaluating cross-lingual and adversarial robustness.  Due to space limitations, we report results for the second test set in the main paper and defer the remaining OOD results to Appendix~\ref{app:ood-details}. For the second test set, we simultaneously shift both the generator family and
the data domain. Specifically, we use four unseen model families,
\ul{\textit{Grok-4.1}}, \ul{\textit{Kimi-K2.5}}, \ul{\textit{Mercury-2}}, and
\ul{\textit{Mistral-Medium-3}}, to rewrite 150 human-written paragraphs from each of
three unseen domains: \ul{\textit{Legal}} for European Union (EU) laws~\cite{chalkidis2021multieurlex}, \ul{\textit{Email}} for email messages~\cite{klimt2004enron}, and
\ul{\textit{Complaints}} for customer complaints~\cite{bureau2025consumer}.  Notably, \ul{\textit{Mercury-2}} is a diffusion-based language
model, introducing an architectural shift beyond standard autoregressive LLMs. This setting is therefore particularly challenging for evaluating \textsc{Reader}'s generalization ability.

\smallskip

\noindent \textbf{Baseline detectors}. 
We compare \textsc{Reader} against \textbf{15} existing detectors, covering: zero-shot detectors \ul{\textit{Likelihood}}, \ul{\textit{Entropy}}, \ul{\textit{LogRank}} \cite{gehrmann2019gltr}, \ul{\textit{LRR}}, \ul{\textit{NPR}} \cite{su2023detectllm}, \ul{\textit{DetectGPT}} \cite{mitchell2023detectgpt}, \ul{\textit{Fast-DetectGPT}} \cite{bao2024fastdetectgpt}, \ul{\textit{Binoculars}} \cite{hans2024spotting},  \ul{\textit{DNAGPT}} \cite{yang2024dnagpt}; and supervised detectors \ul{\textit{RoBERTaBase}}, \ul{\textit{RoBERTaLarge}} \cite{solaiman2019release}, \ul{\textit{RADAR}} \cite{hu2023radar},  \ul{\textit{BiScope}} \cite{guo2024biscope}, \ul{\textit{ImBD}} \cite{chen2025imitate} and \ul{\textit{AdaDetectGPT}} \cite{zhou2025adadetect}.

We also benchmark \textsc{Reader} against \textbf{5} advanced general-purpose LLMs -- \ul{\textit{GPT-5.2}}, \ul{\textit{Claude-Sonnet-4.5}}, \ul{\textit{Qwen3-Max}}, \ul{\textit{Gemini-3-Pro}}, and \ul{\textit{DeepSeek-V3.2}}.

\begin{table}[t]
    \caption{Classification accuracy of various detectors on the \textsc{Read} test set. For each baseline, we report performance under two oracle threshold configurations: global (tuned on the full test set) and per-domain (tuned within each domain). For methods requiring a surrogate LLM, we report results using both Gemma-2-9B(-IT) and Qwen2.5-1.5B-Instruct. The best performance in each column is \textbf{bolded}, and the second-best is \underline{underlined}.}
    \label{tab:oracle_threshold_baselines_best2}
    \begin{center}
        \begin{small}
          \begin{sc}
            \setlength{\tabcolsep}{2.5pt}
            \begin{tabular}{lcccc}
            \toprule
            \textbf{Method} & \multicolumn{2}{c}{\textbf{Global}} & \multicolumn{2}{c}{\textbf{Per-Domain}} \\
            \cmidrule(lr){2-3}\cmidrule(lr){4-5}
            & \textbf{Gemma} & \textbf{Qwen} & \textbf{Gemma} & \textbf{Qwen} \\
            \midrule
            \quad Likelihood        & 0.642 & 0.645 & 0.836 & 0.839 \\
            \quad Entropy           & 0.621 & 0.616 & 0.811 & 0.801 \\
            \quad LogRank           & 0.640 & 0.643 & 0.836 & 0.840 \\
            \quad LRR               & 0.634 & 0.620 & 0.785 & 0.823 \\
            \quad NPR               & 0.619 & 0.611 & 0.812 & 0.813 \\
            \quad DNAGPT            & 0.603 & 0.639 & 0.770 & 0.803 \\
            \quad DetectGPT         & 0.612 & 0.607 & 0.803 & 0.800 \\
            \quad Binoculars        & 0.720 & 0.610 & 0.847 & 0.813 \\
            \quad BiScope           & \underline{0.832} & \underline{0.835} & 0.888 & \underline{0.892} \\
            \quad ImBD              & 0.728 & 0.677 & \underline{0.920} & 0.888 \\
            \quad Fast-DetectGPT    & 0.721 & 0.616 & 0.855 & 0.825 \\
            \quad AdaDetectGPT      & 0.585 & 0.550 & 0.800 & 0.796 \\
            \midrule
            \quad RoBERTa-base      & \multicolumn{2}{c}{0.596} & \multicolumn{2}{c}{0.818} \\
            \quad RoBERTa-large     & \multicolumn{2}{c}{0.568} & \multicolumn{2}{c}{0.798} \\
            \quad RADAR             & \multicolumn{2}{c}{0.588} & \multicolumn{2}{c}{0.779} \\
            \midrule
            \quad \textsc{Reader}   & \multicolumn{4}{c}{\textbf{0.953}} \\
            \bottomrule
        \end{tabular}
      \end{sc}
    \end{small}
  \end{center}
  \vskip -0.1in
\end{table}

\begin{table*}[!t]
\caption{Out-of-distribution detection accuracy across \textbf{three unseen domains} and \textbf{four unseen model families}. Baselines are evaluated under the per-domain configuration with Falcon-7B/Falcon-7B-Instruct employed as the surrogate LLM. The best performance in each column is \textbf{bolded}, and the second-best is \underline{underlined}.}
\label{tab:blackbox_realworld}
\begin{center}
\begin{small}
\setlength{\tabcolsep}{2.6pt}
\renewcommand{\arraystretch}{1.05}
\resizebox{\textwidth}{!}{
\begin{sc}
\begin{tabular}{l|ccc:c|ccc:c|ccc:c|ccc:c}
\toprule
\multicolumn{1}{c}{\textbf{Source LLM}} &
\multicolumn{4}{c}{\textbf{Grok-4.1}} &
\multicolumn{4}{c}{\textbf{Kimi-K2.5}} &
\multicolumn{4}{c}{\textbf{Mercury-2}} &
\multicolumn{4}{c}{\textbf{Mistral-medium-3}} \\
\cmidrule(lr){1-1}\cmidrule(lr){2-5}\cmidrule(lr){6-9}\cmidrule(lr){10-13}\cmidrule(lr){14-17}
\textbf{Method / Domain}
& \textbf{Legal} & \textbf{Email} & \textbf{Complaints} & \textbf{Avg.}
& \textbf{Legal} & \textbf{Email} & \textbf{Complaints} & \textbf{Avg.}
& \textbf{Legal} & \textbf{Email} & \textbf{Complaints} & \textbf{Avg.}
& \textbf{Legal} & \textbf{Email} & \textbf{Complaints} & \textbf{Avg.} \\
\midrule
RoBERTaBase    & 0.577 & 0.560 & 0.530 & 0.556 & 0.580 & 0.577 & 0.537 & 0.564 & 0.630 & 0.587 & 0.563 & 0.593 & 0.627 & 0.597 & 0.597 & 0.607 \\
RoBERTaLarge   & 0.707 & 0.607 & 0.523 & 0.612 & 0.710 & 0.663 & 0.537 & 0.637 & 0.710 & 0.617 & 0.647 & 0.658 & 0.730 & 0.637 & 0.537 & 0.634 \\
Likelihood     & 0.813 & 0.590 & 0.653 & 0.686 & 0.857 & 0.587 & 0.673 & 0.706 & 0.743 & 0.540 & 0.657 & 0.647 & 0.633 & 0.643 & 0.720 & 0.666 \\
Entropy        & 0.687 & 0.553 & 0.703 & 0.648 & 0.737 & 0.530 & 0.747 & 0.671 & 0.667 & 0.563 & 0.697 & 0.642 & 0.663 & 0.637 & 0.763 & 0.688 \\
LogRank        & 0.833 & 0.610 & 0.620 & 0.688 & 0.870 & 0.600 & 0.633 & 0.701 & 0.750 & 0.537 & 0.633 & 0.640 & 0.643 & 0.637 & 0.677 & 0.652 \\
LRR            & \underline{0.847} & 0.660 & 0.550 & 0.686 & 0.840 & 0.633 & 0.550 & 0.674 & 0.700 & 0.530 & 0.550 & 0.593 & 0.683 & 0.567 & 0.567 & 0.606 \\
NPR            & 0.750 & 0.583 & 0.583 & 0.639 & 0.733 & 0.597 & 0.550 & 0.627 & 0.647 & 0.523 & 0.557 & 0.576 & 0.617 & 0.613 & 0.607 & 0.612 \\
DNAGPT         & 0.690 & 0.563 & 0.610 & 0.621 & 0.697 & 0.577 & 0.617 & 0.630 & 0.530 & 0.553 & 0.623 & 0.569 & 0.633 & 0.677 & 0.710 & 0.673 \\
DetectGPT      & 0.720 & 0.557 & 0.603 & 0.627 & 0.743 & 0.597 & 0.563 & 0.634 & 0.610 & 0.553 & 0.577 & 0.580 & 0.557 & 0.603 & 0.617 & 0.592 \\
Binoculars     & 0.837 & 0.667 & 0.537 & 0.680 & 0.873 & 0.697 & 0.540 & 0.703 & 0.773 & 0.630 & 0.543 & 0.649 & 0.587 & 0.613 & 0.600 & 0.600 \\
RADAR          & 0.693 & 0.570 & 0.553 & 0.606 & 0.630 & 0.563 & 0.613 & 0.602 & 0.553 & 0.563 & 0.637 & 0.584 & 0.670 & 0.603 & 0.583 & 0.619 \\
BiScope$^{*}$        & 0.770 & \underline{0.803} & \underline{0.713} & \underline{0.762} & 0.783 & \textbf{0.853} & \underline{0.767} & \underline{0.801} & 0.790 & \underline{0.830} & \underline{0.720} & \underline{0.780} & \underline{0.970} & \textbf{0.860} & \underline{0.797} & \underline{0.876} \\
ImBD$^{*}$           & 0.823 & 0.660 & 0.557 & 0.680 & 0.813 & 0.587 & 0.567 & 0.656 & \underline{0.797} & 0.603 & 0.540 & 0.647 & 0.793 & 0.553 & 0.643 & 0.663 \\
Fast-DetectGPT & \underline{0.847} & 0.683 & 0.537 & 0.689 & \underline{0.880} & 0.713 & 0.527 & 0.707 & 0.780 & 0.627 & 0.540 & 0.649 & 0.590 & 0.623 & 0.603 & 0.606 \\
AdaDetectGPT$^{*}$   & 0.700 & 0.617 & 0.627 & 0.648 & 0.747 & 0.643 & 0.600 & 0.663 & 0.690 & 0.593 & 0.633 & 0.639 & 0.620 & 0.540 & 0.643 & 0.601 \\
\textbf{Reader} & \textbf{0.990} & \textbf{0.853} & \textbf{0.953} & \textbf{0.932} & \textbf{0.987} & \underline{0.850} & \textbf{0.960} & \textbf{0.932} & \textbf{0.957} & \textbf{0.850} & \textbf{0.923} & \textbf{0.910} & \textbf{0.990} & \underline{0.853} & \textbf{0.967} & \textbf{0.937} \\
\bottomrule
\end{tabular}
\end{sc}
}
\end{small}
\end{center}
\vskip -0.1in
\end{table*}

\subsection{In-distribution Evaluation}\label{subsec:indis}
For supervised baseline detectors requiring external training, we train them on the same \textsc{Read} training split as \textsc{Reader} and evaluate all methods together with \textsc{Reader} on the held-out \textsc{Read} test benchmark. Since the test benchmark is balanced between human- and AI-authored texts, we use accuracy as the primary evaluation metric in the main paper and provide the corresponding false positive rates (FPRs) and true positive rates (TPRs) in Tables~\ref{tab:fpr_tpr_read_global} and \ref{tab:fpr_tpr_read_domain} of Appendix~\ref{app:extended-read-results}.
\begin{figure}[t]
    \centering
    \includegraphics[width=1\linewidth]{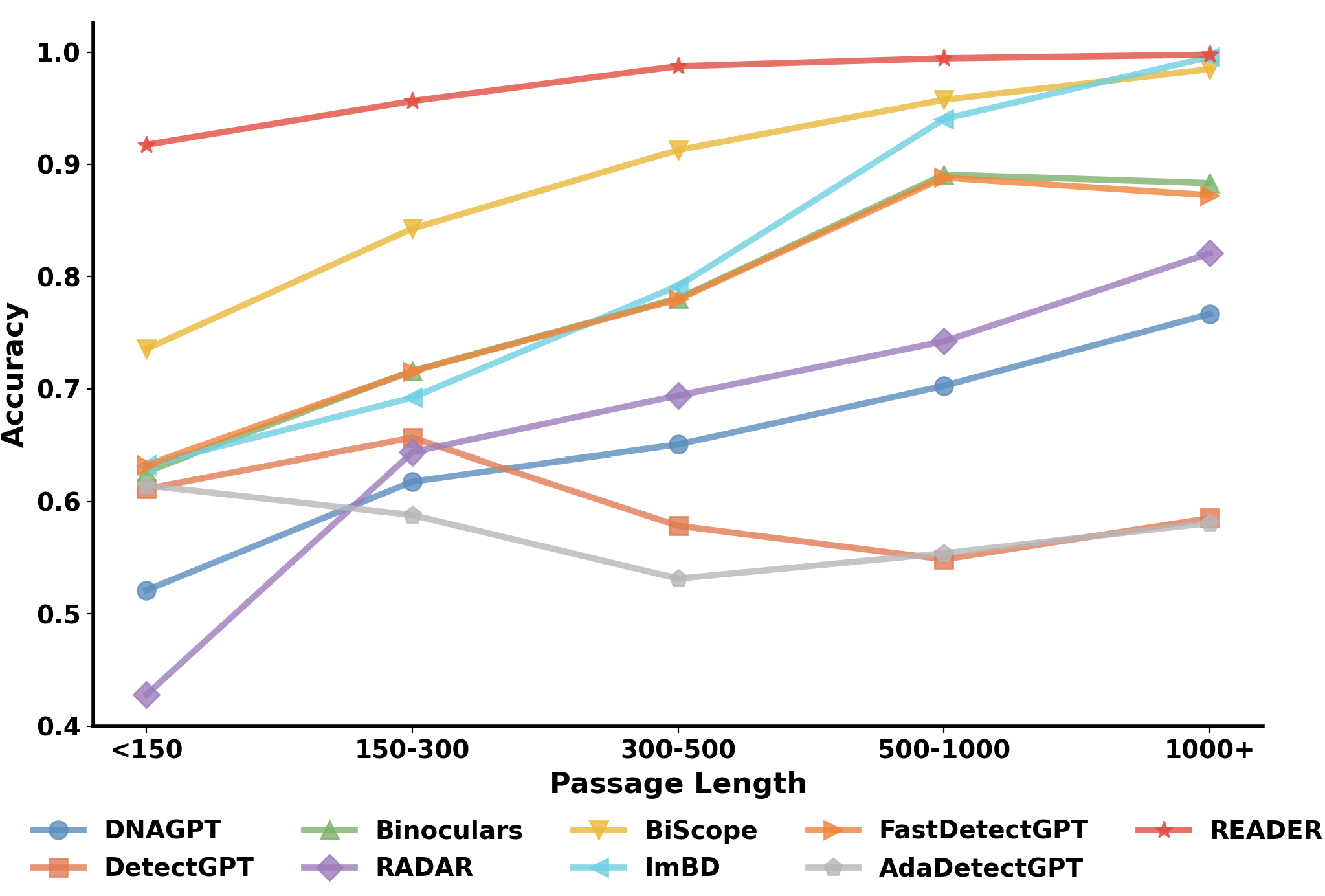}
    \caption{Classification accuracy of various detectors as a function of passage length (measured by the number of tokens).}
    \label{fig:prompt_len_acc}
\end{figure}

\smallskip

\noindent \textbf{Robustness to classification thresholds}. 
This section compares \textsc{Reader} against 15 existing detectors using the \textsc{Read} test set. We first observe that most existing detectors compare a statistical measure or probability score against a pre-defined classification threshold to produce a binary label, and their detection performance is often highly sensitive to this threshold. In contrast, \textsc{Reader} is threshold-agnostic by design. To mitigate the sensitivity of existing detectors to threshold selection, we choose the ``oracle'' threshold for each baseline detector that maximizes their accuracy on the \textsc{Read} test set. Specifically, we consider two oracle configurations: (i) a global configuration that utilizes a single threshold tuned across the entire test set, and (ii) a per-domain configuration where thresholds are optimized independently for each domain. Note that both configurations are infeasible in practice as they require access to test-set labels. They thus provide the best-achievable performance of existing detectors. 

Table~\ref{tab:oracle_threshold_baselines_best2} reports the classification accuracy of \textsc{Reader} alongside baseline detectors under these two oracle configurations. We make the following observations: (i) \textsc{Reader} is highly effective, achieving an accuracy of 95.3\% and outperforming baselines by \textbf{20}-\textbf{40} absolute points under the global configuration and \textbf{5}-\textbf{15} points under the per-domain configuration. The performance gain would be even more substantial compared to baselines using arbitrarily chosen thresholds. (ii) Most existing baselines are largely sensitive to the classification threshold, suggested by their drop in accuracy when moving from per-domain to global configurations. This sensitivity is further highlighted by the empirical results in Table~\ref{tab:read_gemma} of Appendix~\ref{app:extended-read-results}, where the accuracy of several baselines falls below \textbf{50}\% in specific domains under the global configuration. 
(iii) For detectors requiring a surrogate LLM, we report results using both Gemma-2-9B(-IT), following~\citet{zhou2025adadetect,zhou2026learntodistance}, where it has been shown to be a strong surrogate model for detection, and Qwen2.5-1.5B-Instruct, the backbone of \textsc{Reader}. As shown in Table~\ref{tab:oracle_threshold_baselines_best2}, most detectors achieve consistent performance across both models, whereas Fast-DetectGPT suffers a substantial performance drop when switching from Gemma to Qwen under the global configuration.

\smallskip

\noindent \textbf{Robustness to passage lengths}. Prior work has shown that many existing detectors are sensitive to the length of the input text, with longer passages generally being easier to detect \citep{bao2024fastdetectgpt}. To examine this effect, Figure~\ref{fig:prompt_len_acc} reports detection accuracy for \textsc{Reader} and the eight strongest baselines, grouped by passage length. It can be seen that the performance of most baselines is highly dependent on the passage length: their classification accuracy is relatively low for short passages and improves substantially as length increases (e.g., ImBD, Binoculars, and Fast-DetectGPT). 
In contrast, \textsc{Reader} demonstrates consistently strong performance across all lengths, achieving over 90\% accuracy even for passages shorter than 150 tokens, with accuracy approaching 100\% for longer passages. These results indicate that \textsc{Reader} is not only effective but also robust to the length of the input text, remaining reliable for short texts.

\subsection{Out-of-distribution Evaluation}\label{subsec:outofdis}
Table~\ref{tab:blackbox_realworld} reports the classification accuracy of various detectors on the cross-family and cross-domain OOD test dataset, where thresholds for the baseline methods are tuned using the per-domain configuration and the surrogate LLM is set to Falcon following~\citet{hans2024spotting} and~\citet{bao2024fastdetectgpt}. The classification accuracy of baseline detectors using Qwen or Gemma as the surrogate model is reported in  Tables~\ref{tab:blackbox_gemma} and \ref{tab:blackbox_qwen} of Appendix~\ref{app:ood-details}. 

For supervised baselines requiring external training (BiScope, ImBD, and AdaDetectGPT), we additionally consider a target-adapted setting to mitigate their distribution shift and provide a favorable comparison for these methods. Specifically, for each target LLM, we train these detectors on OOD data generated by the same LLM across domains and evaluate them on the held-out domain. This protocol gives these baselines access to target-generator data, and their results are denoted by an asterisk ($^{*}$) in Table~\ref{tab:blackbox_realworld} of the main paper and Table~\ref{tab:blackbox_gemma} of Appendix~\ref{app:ood-details}. Appendix~\ref{app:ood-details} additionally reports the variant, denoted without an asterisk, where these detectors are trained only on \textsc{Read} and evaluated directly on the OOD benchmark.

We make two observations: (i) \textsc{Reader}, despite being trained exclusively on \textsc{Read}, outperforms supervised detectors trained on the OOD data in the majority of cases. This demonstrates the exceptional generalization ability of \textsc{Reader} in detecting AIGT from both unseen model families and unseen domains. (ii)  Since Mercury-2 follows a diffusion-based language-generation paradigm, it introduces a distribution shift not only at the level of model family or training data, but also at the more fundamental level of the language generation mechanism. Under this shift, many existing detectors degrade markedly, suggesting that their decision rules are partly tied to the artifacts of autoregressive decoding. By contrast, \textsc{Reader} remains highly stable, attaining an average accuracy of {\bf 0.91} and improving over the strongest competing baseline by {\bf 13} points. This provides evidence that the robustness of \textsc{Reader} is not confined to conventional autoregressive LLMs, but extends to a substantially different class of generative models.

\begin{table}[!t]
  \caption{Detection results of \textsc{Reader} and five high-capacity LLMs. Acc denotes accuracy. FER denotes the format error rate, defined as the percentage of instances in which the response does not conform to the required  format. UAR denotes the unusable answer rate, defined as the percentage of instances in which the response does not contain a final answer. Best is \textbf{bold}; second-best is \underline{underlined}.}
  \label{tab:accuracy_format_unusable_rates}
  \begin{center}
    \begin{small}
      \begin{sc}
        \begin{tabular}{lccc}
          \toprule
          \textbf{Model} & \textbf{Acc} & \textbf{FER} & \textbf{UAR} \\
          \midrule
          Qwen3-Max         & 0.530 & 0.007 & 0.005 \\
          GPT-5.2           & 0.664 & \textbf{0.000} & \textbf{0.000} \\
          Claude-Sonnet-4.5 & 0.711 & 0.002 & \underline{0.001} \\
          DeepSeek-V3.2     & 0.596 & \underline{0.001} & \underline{0.001} \\
          Gemini-3-Pro      & \underline{0.731} & 0.028 & 0.022 \\
          \textbf{Reader}   & \textbf{0.953} & \textbf{0.000} & \textbf{0.000} \\
          \bottomrule
        \end{tabular}
      \end{sc}
    \end{small}
  \end{center}
  \vskip -0.1in
\end{table}

\subsection{Benchmarking Against General-Purpose LLMs}\label{subsec:generalLLM}
In this section, we compare \textsc{Reader} against five state-of-the-art LLMs on the \textsc{Read} test dataset and report the results in Table \ref{tab:accuracy_format_unusable_rates}\footnote{Due to the high cost of API calls, we evaluate these models on a randomly sampled, representative subset of the \textsc{Read} test set.}. All LLMs receive the same detection prompt (see Prompt~\ref{app:prompt-balanced}) to ensure a fair comparison. 

It can be seen from Table~\ref{tab:accuracy_format_unusable_rates} that: (i) Only \textsc{Reader} and GPT-5.2 consistently return valid outputs with no formatting errors or missing final answers. (ii) \textsc{Reader} achieves the highest accuracy, outperforming other LLMs by a margin of \textbf{20}--\textbf{40}\%. This suggests that with the proposed \textsc{Reader} pipeline, a substantially small model (1.5B) can outperform commercial LLMs that are \textbf{100} or \textbf{1000} larger in scale for the task of AIGT detection. More detailed results and model outputs are provided in Appendix~\ref{sample_analysis}, Tables~\ref{tab:sample_claude_haiku_4_5_1134}--\ref{tab:sample_human_280}.

\begin{table}[t]
\centering
\footnotesize
\renewcommand{\arraystretch}{1.2}
\caption{Answers from \textsc{Reader} and GPT-5.2 on detecting an example human-written text.}
\label{tab:comparison-example}
\begin{tabularx}{\columnwidth}{>{\raggedright\arraybackslash}X}
\toprule

\textbf{Original Text:} Dallas loved to travel. One day he decided to take a vacation to Florida. He swam in the ocean and laid in the sand on the beach. He went deep sea fishing and parasailing. Dallas felt very well rested when he returned from his vacation. \\
\midrule

\rowcolor{headerblue!6}
\multicolumn{1}{c}{\textbf{Model Predictions and Rationales}} \\
\midrule

\rowcolor{lightgray}
\begin{minipage}[t]{\linewidth}
\raisebox{-0.2\height}{\includegraphics[height=1em]{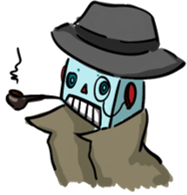}} \textbf{\textsc{Reader}}
\hfill \HUMANbadge\\[-0.15ex]
\vspace{0.25em}
{\color{black!88}
First, the passage uses very simple, colloquial sentence lengths and a personal voice (``Dallas loved to travel'') typical of human casual writing. Second, it shows small stylistic quirks (e.g., ``laid'' for ``lay'') and lacks common AI markers like excessive hedging, mechanical transitions, or over-elaboration. Moreover, the narrative is narrowly focused and conversational rather than generically informative, suggesting human spontaneity. Based on these observations I judge it HUMAN.}
\end{minipage}
\\

\begin{minipage}[t]{\linewidth}
\raisebox{-0.2\height}{\includegraphics[height=1em]{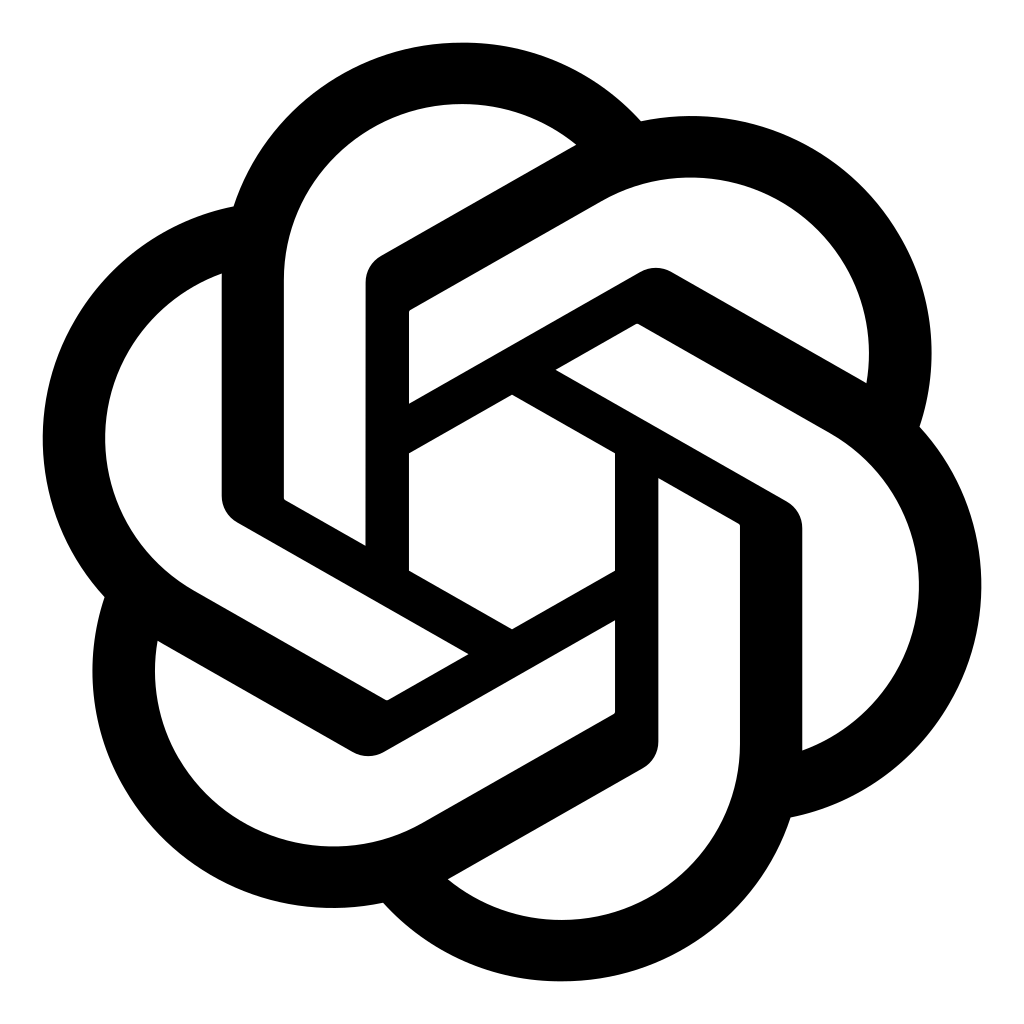}} \textbf{GPT-5.2}
\hfill \AIbadge\\[-0.15ex]
\vspace{0.25em}
{\color{black!88}
First, the text uses very simple, generic sentences and a repetitive subject (``He\ldots\ He\ldots\ He\ldots'') that reads like a template summary rather than a naturally varied human narrative. Second, it follows a tidy, linear sequence of events with no specific sensory details, dialogue, or idiosyncratic wording, which is common in AI-generated or prompted beginner-style outputs. Moreover, the content includes broad, high-level activities (ocean, beach, deep sea fishing, parasailing) presented as a checklist without personal nuance. Therefore, the overall uniformity and bland specificity suggest it is AI-generated.}
\end{minipage}
\\
\bottomrule
\end{tabularx}
\end{table}

\section{Discussion}
This paper introduces \textsc{Reader}, a reasoning-enhanced AI-text detector. Our primary contributions include the development of the \textsc{Read} supervision dataset and the \textsc{Reader} training pipeline. Together, these enable a compact LLM to generate explainable rationales alongside accurate classification labels. Our experimental results demonstrate that \textsc{Reader} outperforms existing supervised and zero-shot detectors in both in-distribution (Section~\ref{subsec:indis}) and out-of-distribution (Section~\ref{subsec:outofdis}) scenarios. Furthermore, \textsc{Reader} outperforms prompted high-capacity LLMs in AIGT detection (Section~\ref{subsec:generalLLM}), and we find that CoT inference improves detection accuracy (Appendix~\ref{sec:ablation}), 
and the generated rationales generally support with the model's final verdicts (Appendix~\ref{app:rationale_coupling}). 
These findings are consistent with prior work showing that explicit reasoning can improve LLM reliability and task performance \cite{wei2022chain,plaat2024reasoning}.

\section*{Limitations}
Despite its strong empirical performance, \textsc{Reader} has a few limitations. First, our evaluation is restricted to text-only inputs, whereas real-world documents often contain non-textual content such as images \citep{wang2023dire,huang2025sida,qi2026difference}. Second, we focus on detecting whether a text is entirely human-authored or AI-authored. In practice, however, mixed authorship frequently occurs in professional writing, where the more relevant question is to localize human- and AI-authored segments rather than assign a single binary label to the entire text \citep{li2024magemachinegeneratedtextdetection}. We leave these directions to future work.

\section*{Ethical Considerations}
\textsc{Reader} is an interpretable detector for AI-generated text, providing both classification labels and human-readable rationales. It can help safeguard against misuse of LLMs in misinformation dissemination, academic dishonesty, and erosion of trust in written communication. In education, it may assist instructors in identifying AI-assisted submissions with explainable evidence; in content moderation, it could help preserve the authenticity of public discourse. However, false positives could lead to unjust accusations against human authors with atypical writing styles, and over-reliance on automated detection may cause inappropriate actions in high-stakes settings. We emphasize that \textsc{Reader} should serve as a supportive tool for human decision-making rather than a sole arbiter of authorship.

\bibliography{reference}

\newpage

\appendix
\newpage

\section{Related Work} \label{Related_work}
AIGT detection aims to determine whether a passage was authored by a human or an LLM. Recent surveys provide a broad view of this rapidly evolving area \citep[e.g.,][]{ji2025overview, 
wu2025survey, yang2024survey, zou2025surveyaigeneratedmediadetection}. Their complementary theoretical perspectives and the inherent limits of the problem have been discussed by \citet{chakraborty2024position} and \citet{sadasivan2025can}. Existing methods can be grouped into three families: (i) supervised (ML-based) detectors trained on labeled human/LLM corpora, (ii) training-free (often called zero-shot) detectors that compute a statistic on a text or under a reference model, and (iii) watermarking methods that embed a detectable signal during generation. The first two approaches are passive, as they classify content based solely on the text without any control over the generator, whereas watermarking is active, as it requires control over the generation process. 

\ul{\textit{Our work focuses on passive detection, and specifically on detectors that provide human-readable rationales in addition to a binary label}}. We first review the first two families of methods, followed by a discussion of explainable detection approaches.

\smallskip

\noindent \textbf{Supervised (ML-based) detectors.}
Supervised detectors cast AIGT detection as binary classification using labeled human- and machine-authored text. A large class of approaches relies on engineered or LLM-assisted features, including lexical statistics such as TF--IDF and $n$-grams \citep[e.g.,][]{solaiman2019release}, token-rank/token-probability statistics \citep[e.g.,][]{verma2024ghostbuster}, likelihood or cross-entropy loss evaluated under a surrogate LLM  \citep[e.g.,][]{guo2024biscope}, and rewrite- or paraphrase-consistency signals that measure the difference between an input text and its LLM-generated rewrite \citep[e.g.,][]{mao2024raidar, huang2025magret, park2025dart}. 

Other methods extract latent representations and train a classifier in embedding space \citep[e.g.,][]{yu2024textflu}, learn contrastive features for generalization \citep[e.g.,][]{guo2024detective}, or combine multiple detectors via ensembling \citep[e.g.,][]{abburi2023generative}. 

Another line of work directly fine-tunes pretrained encoders (e.g., BERT/RoBERTa) as detectors  \citep[e.g.,][]{solaiman2019release, guo2023close,ippolito2020bert}, or leverages reward models from reinforcement learning from human feedback \citep{lee2024remodetect}. 

These approaches have been extended to specifically target robustness to paraphrasing and adversarial attacks \citep[e.g.,][]{hu2023radar, koike2024outfox, sadasivan2025can, krishna2023paraphrasing}, short-text settings \citep[e.g.,][]{tian2024multiscale, kumarage2023stylometric}, bias and fairness concerns \citep[e.g.,][]{liang2023gpt}, domain-specific deployment \citep[e.g.,][]{herbold2023large, liao2023differentiating} and black-box settings   \citep[e.g.,][]{zeng2024dlad, yu2024dpic}. Relatedly, \citet{chen2025imitate} study the detection of machine-revised text aligning to machine stylistic preferences. 

\smallskip

\noindent \textbf{Zero-shot (Training-free) detectors.}
These approaches compute a hand-crafted score on the input text and identify its provenance by comparing the score against a predetermined threshold, without using labeled data. A prominent family computes the score based on token-probability statistics \citep[e.g.,][]{gehrmann2019gltr,su2023detectllm}, and has been extended with perturbation-based methods that probe the stability of model likelihood under edits \citep[e.g.,][]{mitchell2023detectgpt, bao2024fastdetectgpt, su2023detectllm, hans2024spotting, xu2025trainingfree, zhou2025adadetect,li2026segmenting}. 

Another widely used approach compares the input to an LLM rewrite or regeneration and sets the detection score to the edit distance or semantic divergence \citep{zhu2023beat, nguyen2024simllm, yang2024dnagpt, sun2025zero, hao2025learning}. 

Other training-free scores are computed based on intrinsic dimensionality estimates  \citep[e.g.,][]{tulchinskii2023intrinsic}, hidden representation patterns \citep[e.g.,][]{chen2025repreguard}, and distributional two-sample tests such as maximum mean discrepancy \citep[e.g.,][]{zhang2024detecting, song2025deep}.  

Finally, several works explore using prompted LLMs themselves as detectors  \citep[e.g.,][]{bhattacharjee2023fightingfirechatgptdetect, abburi-etal-2024-deloitte}.

\smallskip

\noindent \textbf{Explainable detectors}.
Most detectors output only a label or a score, which can be insufficient in high-stakes settings where users need to understand \emph{why} a decision was made. Prior work on explainable detection often uses post-hoc attribution tools such as LIME or SHAP \citep[e.g.,][]{mitrovic2023chatgpt, joshi2024hullmihumanvsllm, yuan2025emmm}, or visualization systems such as GLTR \citep[e.g.,][]{gehrmann2019gltr}. However, token-level attributions can be difficult to interpret and may not reflect the underlying decision process. \citet{ji2025detectingmachinegeneratedtextsjust} analyze the challenges of natural-language justifications in AIGT detection, but do not provide a scalable detector trained to produce explanations. Recent work on AI image detection explores explainability via multimodal LLMs  \citep[e.g.,][]{liu2024forgerygpt, xu2024fakeshield, huang2025sida, lin2025detectingmultimediageneratedlarge}. We bring a similar perspective to text detection: our model produces explicit reasoning traces, offering both strong detection performance and informative natural-language explanations.

\section{\textsc{Read} Data Construction}

\subsection{Prompt Selection for Data Construction}\label{prompts}
We use GPT-5 as the teacher model to generate structured, human-readable
rationales that accompany authorship labels and provide explicit linguistic
evidence for the prediction.  Since the quality of the resulting rationales can
depend substantially on the prompt used to query the teacher model, we perform
prompt-level quality control before constructing the final rationale-augmented
training data.  This procedure is designed to select a stable and informative
teacher prompt.

Specifically, we first design a set of candidate teacher prompts for
GPT-5, covering different levels of rationale supervision and instruction
detail.  We manually inspect these prompts to remove prompts
that frequently produce invalid JSON, unstable verdicts, overly generic
explanations, weak grounding in the input text, or rationales that do not
support the predicted label.  This initial prompt-level screening retains four representative prompting strategies: \textit{Balanced}, 
\textit{Concise}, \textit{Rubric-Based}, and \textit{One-Shot}.  These prompts
differ in how concise or detailed
the rationale should be, whether linguistic cues are specified through a
rubric, and whether in-context examples are provided.

To further reduce dependence on the same model that generates the rationales,
we evaluate the retained candidate prompts using an independent model-based quality assessment.  Following the LLM-as-a-judge protocol of \citet{gu2024survey}, we use Qwen3-Max as the judge.  For
each candidate prompt, Qwen3-Max scores the generated rationales along three
dimensions: \emph{specificity}, \emph{grounding}, and \emph{coherence}.
Specificity measures whether the rationale provides concrete and non-generic evidence.  Grounding measures whether the explanation is supported by the input
text.  Coherence measures whether the rationale is logically organized and
consistent with the final prediction.  We additionally report the corresponding classification accuracy and an overall quality score that combines rationale quality with prediction correctness.

Table~\ref{tab:rationale_prompt_filtering} reports the prompt-selection results.
Among the GPT-5 candidate prompts, the \textit{Balanced} prompt obtains the
highest overall quality score.  We therefore use this prompt to generate the
final rationales for constructing the rationale-augmented training set used by \textsc{Reader}.  The exact prompt templates are listed below as Prompt~\ref{app:prompt-balanced} (Balanced), Prompt~\ref{app:prompt-noncot} (Non-CoT), Prompt~\ref{app:prompt-concise} (Concise), Prompt~\ref{app:prompt-rubric} (Rubric-Based), and Prompt~\ref{app:prompt-oneshot} (One-Shot). The Non-CoT template asks only for the final verdict without an explicit rationale and is used for the ablation analysis in Appendix~\ref{sec:ablation}.
\begin{table}[t]
\centering
\caption{\textsc{Read} filtering}
\label{tab:error_analysis}
\begin{small}
\begin{tabular}{lc}
\toprule
\textbf{Category} & \textbf{Rate} \\
\midrule
Correct predictions & 71.88\% \\
Wrong predictions & 27.82\% \\
Parse errors & 0.30\% \\
\bottomrule
\end{tabular}
\end{small}
\end{table}
\begin{table*}[p]

\centering

\small

\caption{Prompt-selection results for GPT-5 rationale generation. Specificity, grounding, and coherence are scored on a 1--5 scale by an independent Qwen3-Max judge. Accuracy denotes the corresponding detection accuracy. The overall quality score combines rationale-quality scores with classification accuracy. The \textit{Non-CoT} prompt does not generate rationales and is included as a non-rationale baseline.}

\label{tab:rationale_prompt_filtering}

\resizebox{\linewidth}{!}{

\begin{tabular}{lccccc}

\toprule

\textbf{Dimension} & \textbf{Balanced} & \textbf{Concise} & \textbf{Rubric-Based} & \textbf{One-Shot} & \textbf{Non-CoT} \\

\midrule

Specificity (out of 5)      & 4.18 & 4.10 & 4.39 & 4.46 & N/A \\

Grounding (out of 5)        & 4.88 & 4.58 & 4.96 & 4.83 & N/A \\

Coherence (out of 5)        & 4.83 & 4.62 & 4.68 & 4.65 & N/A \\

Accuracy                    & 72\% & 65\% & 63\% & 66\% & 62\% \\

Overall quality (out of 5)  & 4.38 & 4.18 & 4.20 & 4.18 & N/A \\

\bottomrule

\end{tabular}

}

\end{table*}

\subsection{Reasoning Data Filtering}\label{app:filtering}
We use Prompt~\ref{app:prompt-balanced} to collect reasoning traces from GPT-5. Table~\ref{tab:error_analysis} reports the filtering proportions and separates three outcomes: correct predictions retained as candidate rationale supervision, wrong predictions discarded because the teacher verdict conflicts with the ground-truth authorship label, and parse errors discarded because the output cannot be reliably converted into the required structured format. The retained example in Figure~\ref{fig:filter-correct-human} illustrates the desired case: the verdict is correct and the rationale cites text-specific evidence, including idiosyncratic details, informal punctuation, and lived-in complaints. The wrong-prediction example in Figure~\ref{fig:filter-wrong-ai} shows why label agreement is necessary even when the rationale sounds plausible; the teacher incorrectly treats an AI-generated review-like passage as human-written. Figures~\ref{fig:filter-incomplete-string}, \ref{fig:filter-unescaped-quote}, \ref{fig:filter-empty-output}, and \ref{fig:filter-extra-closing} show representative parse or format failures, including incomplete strings, unescaped quotes, empty outputs, and malformed extra closing symbols. These examples motivate the two-stage filter used for \textsc{Read}: we first require a correct teacher decision, then require a usable and internally consistent rationale before adding the sample to the SFT rationale set.

\begin{figure*}[p]
\centering
\refstepcounter{prompttemplate}\label{app:prompt-balanced}
\begin{tcolorbox}[
  colback=gray!3,
  colframe=blue!40,
  title={Prompt~\theprompttemplate: Balanced Prompt},
  enhanced,
  boxrule=0.6pt,
  arc=1mm,
  width=\textwidth,
  left=1.2mm,right=1.2mm,top=1.2mm,bottom=1.2mm
]
\begin{Verbatim}[
  fontsize=\footnotesize,
  breaklines=true,
  breakanywhere=true,
  formatcom=\raggedright,
  breaksymbolleft={},   
  breaksymbolright={}   
]
You are an expert in textual forensics. Your job is to decide whether the following text is AI-generated or human-written. 
Please reason step by step, considering the style, structure, and content of the text. Justify your decision by pointing out specific features or characteristics of the text that suggest it is either AI or human-written.
Use conjunction words such as "first", "second", "moreover" etc and end with your decision in your reasoning.
Return STRICT JSON with fields exactly:
{
  "rationale": "3-5 concise sentences explaining your reasoning",
  "verdict": "AI" or "HUMAN"
}
\end{Verbatim}
\end{tcolorbox}
\end{figure*}

\begin{figure*}[p]
\centering
\refstepcounter{prompttemplate}\label{app:prompt-noncot}
\begin{tcolorbox}[
  colback=gray!3,
  colframe=blue!40,
  title={Prompt~\theprompttemplate: Non-CoT Prompt},
  enhanced,
  boxrule=0.6pt,
  arc=1mm,
  width=\textwidth,
  left=1.2mm,right=1.2mm,top=1.2mm,bottom=1.2mm
]
\begin{Verbatim}[
  fontsize=\footnotesize,
  breaklines=true,
  breakanywhere=true,
  formatcom=\raggedright,
  breaksymbolleft={},   
  breaksymbolright={}   
]
You are an expert in textual forensics. Decide if the passage was written by an AI system or a human.
Return STRICT JSON only:
{
  "verdict": "AI" or "HUMAN"
}
\end{Verbatim}
\end{tcolorbox}
\end{figure*}

\begin{figure*}[p]
\centering
\refstepcounter{prompttemplate}\label{app:prompt-concise}
\begin{tcolorbox}[
  colback=gray!3,
  colframe=blue!40,
  title={Prompt~\theprompttemplate: Concise Prompt},
  enhanced,
  boxrule=0.6pt,
  arc=1mm,
  width=\textwidth,
  left=1.2mm,right=1.2mm,top=1.2mm,bottom=1.2mm
]
\begin{Verbatim}[
  fontsize=\footnotesize,
  breaklines=true,
  breakanywhere=true,
  formatcom=\raggedright,
  breaksymbolleft={},   
  breaksymbolright={}   
]
You are an expert in textual forensics. Decide if the passage was written by an AI system or a human. Briefly justify your decision by pointing to 2-3 specific cues in the passage.
Return STRICT JSON only: 
{ 
  "rationale": "2-4 short, text-specific sentences",  
  "verdict": "AI" or "HUMAN" 
}
\end{Verbatim}
\end{tcolorbox}
\end{figure*}

\begin{figure*}[p]
\centering
\refstepcounter{prompttemplate}\label{app:prompt-rubric}
\begin{tcolorbox}[
  colback=gray!3,
  colframe=blue!40,
  title={Prompt~\theprompttemplate: Rubric-Based Prompt},
  enhanced,
  boxrule=0.6pt,
  arc=1mm,
  width=\textwidth,
  left=1.2mm,right=1.2mm,top=1.2mm,bottom=1.2mm
]
\begin{Verbatim}[
  fontsize=\footnotesize,
  breaklines=true,
  breakanywhere=true,
  formatcom=\raggedright,
  breaksymbolleft={},   
  breaksymbolright={}  
]
You are an expert in textual forensics. Your goal is decide whether the following text is AI-generated or human-written.
Judge using only stylistic/structural signals in the passage (ignore topic correctness, outside knowledge, or author identity).
AI-leaning cues (examples): overly smooth "editorial" prose with low variance, checklist-like organization (enumerations, parallel clauses, balanced paragraphs), generic claims that stay high-level (few concrete names/numbers/places), careful hedging plus confident wrap-up, consistent polite/helpful tone, redundancy via rephrasing (same point restated with synonyms), stock academic/blog connectors ("moreover", "in summary", "it is important to note"), unnatural completeness (covers many angles with little depth), few genuine constraints or omissions, few surprising choices of detail.
HUMAN-leaning cues (examples): uneven texture (some sharp lines, some rough ones), spontaneous specificity (incidental details, time/place markers, small numbers, proper nouns) that don't feel like "filler," self-repairs and micro-revisions (dashes, parentheticals, "wait/actually"), idiosyncratic rhythm/voice (quirky phrasing, humor, irritation, uncertainty), locally inconsistent structure (digressions, abrupt transitions, emphasis patterns), selective coverage (focuses on what mattered, leaves gaps), concrete causal chains tied to experience ("I did X, then Y happened"), occasional asymmetry or imperfection that still feels intentional.
Return STRICT JSON with fields exactly:
{
  "rationale": "2-5 text-specific sentences; you can cite concrete cues (e.g., structure, tone, specificity) given in the instruction",
  "verdict": "AI" or "HUMAN"
}
\end{Verbatim}
\end{tcolorbox}
\end{figure*}

\begin{figure*}[p]
\centering
\refstepcounter{prompttemplate}\label{app:prompt-oneshot}
\begin{tcolorbox}[
  colback=gray!3,
  colframe=blue!40,
  title={Prompt~\theprompttemplate: One-Shot Prompt},
  enhanced,
  boxrule=0.6pt,
  arc=1mm,
  width=\textwidth,
  left=1.2mm,right=1.2mm,top=1.2mm,bottom=1.2mm
]
\begin{Verbatim}[
  fontsize=\footnotesize,
  breaklines=true,
  breakanywhere=true,
  formatcom=\raggedright,
  breaksymbolleft={},   
  breaksymbolright={}   
]
You are an expert in textual forensics. Your goal is decide whether the following text is AI-generated or human-written.
I will show you two examples of each class, you need to learn to illustrate the differences and generalize to the new examples.
AI example:
"This report details the implementation of AlexNet (Krizhevsky et al., 2012) using Theano and its straightforward data parallelism across multiple GPUs. The performance achieved with 2 GPUs is similar to the state-of-the-art results obtained with the Caffe library (Jia et al., 2014) on a single GPU. As far as we know, this is the first open-source AlexNet implementation written in Python."
HUMAN example:
"In this report, we describe a Theano-based AlexNet (Krizhevsky et al., 2012) implementation and its naive data parallelism on multiple GPUs. Our performance on 2 GPUs is comparable with the state-of-art Caffe library (Jia et al., 2014) run on 1 GPU. To the best of our knowledge, this is the first open-source Python-based AlexNet implementation to-date."
AI-leaning cues (examples): instruction-following scaffolding ("in summary", "moreover", "first/second"), steady unvarying tone, generic but exhaustive coverage lacking concrete specifics, repetition with shallow paraphrase/synonym churn, tidy paragraph symmetry, formulaic openings/closings, confident abstractions without lived detail.
HUMAN-leaning cues (examples): concrete episodic/sensory detail, idiosyncratic asides or self-corrections, mixed register, locally messy structure, specific experiential references, unusual but intentional turns of phrase.
Return STRICT JSON with fields exactly:
{
  "rationale": "2-5 short, text-specific sentences",
  "verdict": "AI" or "HUMAN"
}
\end{Verbatim}
\end{tcolorbox}
\end{figure*}

\FloatBarrier

\begin{figure*}[p]
\centering
\begin{tcolorbox}[
    enhanced,
    colback=blue!5,
    colframe=blue!60,
    coltitle=white,
    fonttitle=\bfseries,
    title={Correct Prediction (Ground Truth: Human)},
    boxed title style={colback=blue!70},
    attach boxed title to top left={yshift=-2mm, xshift=4mm},
    top=3mm,
    arc=1.5mm,
    boxrule=0.6pt,
    width=\textwidth,
    fontupper=\footnotesize
]

\textbf{Original Text:}\\
First, let me say that I'm a bit grumpy about Las Vegas in general. But this is about the hotel at Bally's. We stayed in the Indigo tower. I'm assuming this hasn't been renovated recently. Not sure. It didn't look run down. But the features of the bathroom and the ... friendly. I didn't eat at the restaurants, except for the food court. The price of wi-fi is outrageous considering you have to pay per device. It's not unreasonable that a couple might have two phones and a laptop or tablet. So that would be nearly 40 a day. That's unreasonable.

\vspace{2mm}
\textbf{Prompt:}\\
\ttfamily
You are an expert in textual forensics... decide whether the following text is AI-generated or human-written... Return STRICT JSON with fields exactly: \{ "rationale": "...", "verdict": "AI" or "HUMAN" \}.

\vspace{2mm}
\textbf{Response:}
\begin{Verbatim}[breaklines=true, breaksymbolleft={}, commandchars=~<>]
{\n  "rationale": "First, the text contains conversational, idiosyncratic details (e.g., \'Indigo tower\', moving the bedside table, unplugging the alarm clock, \'BLASTS\' in all caps, \'Ugh.\') and pragmatic inconsistencies that are typical of a human traveler reporting a personal stay. Second, the uneven punctuation, colloquial interjections, small typos (\'Checkin\'), and emotional asides (\'I\'m NOT going to... Ever\', \'Hell, you could...\') suggest spontaneous human writing rather than the more uniform, polished output of an AI. Moreover, the specific, mundane operational complaints (vending machines not taking small bills, outlets placement, per-device Wi‑Fi pricing) and sensory cues (mildew scent, humid lingering) feel lived-in and particular rather than generically generated. Therefore my decision is HUMAN",\n  "verdict": "~codeblue HUMAN~codeblack "\n}
\end{Verbatim}
\end{tcolorbox}
\caption{Retained reasoning-data example with a correct teacher prediction for a human-written passage.}
\label{fig:filter-correct-human}
\end{figure*}

\begin{figure*}[p]
\centering
\begin{tcolorbox}[
    enhanced,
    colback=white,
    colframe=gray!60,
    coltitle=white,
    fonttitle=\bfseries,
    title={Wrong Prediction (Ground Truth: AI)},
    boxed title style={colback=gray!80},
    attach boxed title to top left={yshift=-2mm, xshift=4mm},
    top=3mm,
    arc=1.5mm,
    boxrule=0.6pt,
    width=\textwidth,
    fontupper=\footnotesize
]

\textbf{Original Text:}\\
Ok so to start the food is actually really good and consistent. The reason I rated them lower is because of the customer service. I have been several times and have had some negative interactions with the staff. Could you please provide more information about the interactions you had? I may be able to give you some advice on how to handle the situation in the future.

\vspace{2mm}
\textbf{Prompt:}\\
\ttfamily
You are an expert in textual forensics... decide whether the following text is AI-generated or human-written... Return STRICT JSON with fields exactly: \{ "rationale": "...", "verdict": "AI" or "HUMAN" \}.

\vspace{2mm}
\textbf{Response:}\\
\begin{Verbatim}[breaklines=true, breaksymbolleft={}, commandchars=~<>]
{\n  \"rationale\": \"First, the informal and slightly unpolished phrasing like \\\"Ok so to start\\\" and the redundant \\\"have been several times and have had\\\" mirror natural human speech rather than tightly edited AI output. Second, the text mixes a personal review with a direct follow-up question and offer of advice, showing conversational turn-taking and experiential detail typical of a human writer. Moreover, the brief, context-specific content and uneven tone suggest spontaneous human composition rather than a generated template. HUMAN\",\n  \"verdict\": \"~codered HUMAN"\n~codeblack}
\end{Verbatim}
\end{tcolorbox}
\caption{Filtered reasoning-data example with an incorrect teacher prediction for an AI-generated passage.}
\label{fig:filter-wrong-ai}
\end{figure*}

\begin{figure*}[p]
\centering
\begin{tcolorbox}[
    enhanced,
    colback=white,
    colframe=gray!60,
    coltitle=white,
    fonttitle=\bfseries,
    title={Incomplete String},
    boxed title style={colback=gray!80},
    attach boxed title to top left={yshift=-2mm, xshift=4mm},
    top=3mm,
    arc=1.5mm,
    boxrule=0.6pt,
    width=\textwidth,
    fontupper=\footnotesize
]

\textbf{Original Text:}\\
Fred hadn't bought new clothes in many Year's. His wife and kids wanted him to update his old clothing collection. Finally they convinced him to come shopping. Fred loved it so much he began shopping every weekend. Now his family wishes they had never told him to anything.

\vspace{2mm}
\textbf{Prompt:}\\
\ttfamily
You are an expert in textual forensics... decide whether the following text is AI-generated or human-written... Return STRICT JSON with fields exactly: \{ "rationale": "...", "verdict": "AI" or "HUMAN" \}.
\\
\textbf{Response:}
\begin{Verbatim}[breaklines=true, breaksymbolleft={}, commandchars=~<>]
~codered {\n  \"rationale\": \"First, the text contains multiple informal errors (e.g., the misplaced apostrophe in Year's and the,~codeblack
\end{Verbatim}
\end{tcolorbox}
\caption{Filtered reasoning-data example with an incomplete output string.}
\label{fig:filter-incomplete-string}
\end{figure*}

\begin{figure*}[p]
\centering
\begin{tcolorbox}[
    enhanced,
    colback=white,
    colframe=gray!60,
    coltitle=white,
    fonttitle=\bfseries,
    title={Unescaped Quote},
    boxed title style={colback=gray!80},
    attach boxed title to top left={yshift=-2mm, xshift=4mm},
    top=3mm,
    arc=1.5mm,
    boxrule=0.6pt,
    width=\textwidth,
    fontupper=\footnotesize
]

\textbf{Original Text:}\\
Let me begin by defining what I mean when I say hyper-rationality. Hyper-rationality is the social expectation and pressure that every human action and institution must be teologically justified by rational means, rather than a mixture of reason, emotion, tradition, religion, evolutionary instinct etc. I will use one clear example, ... one, it is more effective to report it than downvote it. Speaking of which, [downvotes don't change views] ! If you are thinking about submitting a CMV yourself, please have a look through our [popular topics wiki] first. Any questions or concerns? Feel free to [message us] . Happy CMVing!

\vspace{2mm}
\textbf{Prompt:}\\
\ttfamily
You are an expert in textual forensics... decide whether the following text is AI-generated or human-written... Return STRICT JSON with fields exactly: \{ "rationale": "...", "verdict": "AI" or "HUMAN" \}.

\vspace{2mm}
\textbf{Response:}\\
\begin{Verbatim}[breaklines=true, breaksymbolleft={}, commandchars=~<>]
{\n  \"rationale\": \"First, the text contains forum-specific artifacts (a moderator footnote and a scrambled subreddit link) and formatting glitches that typically come from a human posting or scraped content. Second, the writing has informal, idiosyncratic phrasing and minor misspellings (e.g., \\\"teology\\\", \\\"off-spring\\\") and uneven structure consistent with a casual Reddit post. Moreover, the anecdotal voice, hedging (~codered \"3-4 arguments\"~codeblack , \\\"shall we say\\\"), and rhetorical asides are characteristic of human writers rather than the smoother, more neutral output of most AIs. Therefore I conclude HUMAN.\",\n  \"verdict\": \"HUMAN\"\n}
\end{Verbatim}
\end{tcolorbox}
\caption{Filtered reasoning-data example with an unescaped quote in the structured output.}
\label{fig:filter-unescaped-quote}
\end{figure*}

\begin{figure*}[p]
\centering
\begin{tcolorbox}[
    enhanced,
    colback=white,
    colframe=gray!60,
    coltitle=white,
    fonttitle=\bfseries,
    title={Empty Output},
    boxed title style={colback=gray!80},
    attach boxed title to top left={yshift=-2mm, xshift=4mm},
    top=3mm,
    arc=1.5mm,
    boxrule=0.6pt,
    width=\textwidth,
    fontupper=\footnotesize
]

\textbf{Original Text:}\\{}
 [step] Rub your palms together to distribute the product over your hands, then massage it evenly through hair. title Comb through your hair with your fingers while gently blow-drying it on a low setting.

\vspace{2mm}
\textbf{Prompt:}\\
\ttfamily
You are an expert in textual forensics... decide whether the following text is AI-generated or human-written... Return STRICT JSON with fields exactly: \{ "rationale": "...", "verdict": "AI" or "HUMAN" \}.

\vspace{2mm}
\textbf{Response:}\\
\ttfamily
\end{tcolorbox}
\caption{Filtered reasoning-data example with an empty teacher output.}
\label{fig:filter-empty-output}
\end{figure*}

\FloatBarrier
\begin{figure*}[!t]
\vspace{-10pt}
\centering
\begin{tcolorbox}[
    enhanced,
    colback=white,
    colframe=gray!60,
    coltitle=white,
    fonttitle=\bfseries,
    title={Extra Closing},
    boxed title style={colback=gray!80},
    attach boxed title to top left={yshift=-2mm, xshift=4mm},
    top=3mm,
    arc=1.5mm,
    boxrule=0.6pt,
    width=\textwidth,
    fontupper=\footnotesize
]

\textbf{Original Text:}\\
This paper analyzes the effects of approximate multiplication when performing inferences on deep convolutional neural networks (CNNs). The approximate multiplication can reduce the cost of underlying circuits so that CNN inferences can be performed more efficiently in hardware accelerators. The study identifies the critical factors in the convolution, fully-connected, and ... comparison of Mitch- w 6 against bfloat16 is presented, where a MAC operation saves up to 80 of energy compared to the bfloat16 arithmetic. The most far-reaching contribution of this paper is the analytical justification that multiplications can be approximated while additions need to be exact in CNN MAC operations.

\vspace{2mm}
\textbf{Prompt:}\\
\ttfamily
You are an expert in textual forensics... decide whether the following text is AI-generated or human-written... Return STRICT JSON with fields exactly: \{ "rationale": "...", "verdict": "AI" or "HUMAN" \}.

\vspace{2mm}
\textbf{Response:}\\
\begin{Verbatim}[breaklines=true, breaksymbolleft={}, commandchars=~<>]
{\n  \"rationale\": \"First, the passage contains stilted and repetitive phrasing with awkward numeric formatting and missing units (e.g., 'Mitch- w 6', 'within 0.2', '80 of energy'), which is characteristic of automated generation. Second, it has vague, high-level claims and inconsistent technical specifics (e.g., 'recognized network architectures', no datasets or baselines) that suggest summarization rather than original human exposition. Moreover, minor grammar, spacing, and hyphenation errors and oddly phrased sentences point to a model paraphrasing source material rather than careful human writing. Finally, these combined artifacts lead me to conclude the text is AI\"~codered },\n  \"verdict\": \"AI\"\n}~codeblack
\end{Verbatim}
\end{tcolorbox}
\caption{Filtered reasoning-data example with a malformed extra closing symbol.}
\label{fig:filter-extra-closing}
\end{figure*}

\section{Implementation Details and Additional Results}
\label{app:additional-results}

\subsection{Implementation Details}\label{app:implementation-details}
For \textsc{Reader}-SFT training, we follow the \textit{TRL: Transformer Reinforcement Learning}\footnote{https://github.com/huggingface/trl} framework \cite{vonwerra2020trl} and train for 3 epochs using the high-quality CoT dataset. For \textsc{Reader}-GRPO, we use the \textit{verl: Volcano Engine Reinforcement Learning for LLMs}\footnote{https://github.com/verl-project/verl} framework \cite{sheng2024hybridflow} and train for 6 epochs using the GRPO data. During inference, we use temperature 0 to greedily obtain almost deterministic responses.

For baseline methods, we use the implementations provided in Fast-DetectGPT\footnote{https://github.com/baoguangsheng/fast-detect-gpt} and AdaDetectGPT\footnote{https://github.com/Mamba413/AdaDetectGPT}, which provide standardized pipelines for the evaluated detectors. Specifically, for the BERT-based classifiers we use the \textit{openai-community/roberta-base-openai-detector} and \textit{openai-community/roberta-large-openai-detector} checkpoints provided by the OpenAI community on HuggingFace, and for RADAR we use the released RADAR checkpoint \textit{TrustSafeAI/RADAR-Vicuna-7B}. For DetectGPT, we use the T5 masking model \textit{t5-3b} to construct perturbations. For DetectLLM, NPR/LRR scores are computed on DetectGPT perturbations, consistent with the repository implementation. All baselines are run following the standard procedures and default configurations provided by these repositories, unless stated otherwise.

\begin{table*}[!t]
\caption{\textsc{Read} benchmark: per-domain classification accuracy of baseline detectors and \textsc{Reader} under the global configuration. For methods requiring a surrogate/reference LLM, we use Gemma-2-9B(-IT). \textbf{Bold} denotes the best performance in each column and \underline{underlined} denotes the second-best.}\label{tab:read_gemma}
\begin{center}
\begin{small}
\begin{sc}
\setlength{\tabcolsep}{2pt}
\renewcommand{\arraystretch}{1.05}
\resizebox{\textwidth}{!}{%
\begin{tabular}{l|cccccccccccc}
\toprule
\textbf{Method / Domain} & \textbf{CMV} & \textbf{ELI5} & \textbf{HellaSwag} & \textbf{ROC} & \textbf{SCI} & \textbf{SQuAD} & \textbf{TLDR} & \textbf{WP} & \textbf{XSum} & \textbf{Yelp} & \textbf{Rewrite} & \textbf{Avg.} \\
\midrule
RoBERTaBase & 0.833 & 0.796 & 0.792 & 0.698 & 0.713 & 0.775 & 0.714 & 0.835 & 0.800 & 0.768 & 0.156 & 0.596 \\
RoBERTaLarge & 0.788 & 0.764 & 0.758 & 0.702 & 0.680 & 0.739 & 0.695 & 0.776 & 0.763 & 0.714 & 0.144 & 0.568 \\
Likelihood & 0.877 & 0.840 & 0.848 & 0.871 & 0.754 & 0.726 & 0.779 & 0.867 & 0.529 & 0.764 & 0.283 & 0.642 \\
Entropy & 0.828 & 0.760 & 0.830 & 0.883 & 0.637 & 0.592 & 0.725 & 0.828 & 0.370 & 0.709 & 0.372 & 0.621 \\
LogRank & 0.877 & 0.837 & 0.850 & 0.862 & 0.754 & 0.724 & 0.773 & 0.865 & 0.540 & 0.768 & 0.275 & 0.640 \\
LRR & 0.763 & 0.746 & 0.761 & 0.752 & 0.572 & 0.677 & 0.644 & 0.828 & 0.580 & 0.695 & 0.451 & 0.634 \\
NPR & 0.615 & 0.665 & 0.838 & 0.897 & 0.664 & 0.711 & 0.752 & 0.472 & 0.398 & 0.684 & 0.497 & 0.619 \\
DNAGPT & 0.803 & 0.779 & 0.612 & 0.545 & 0.727 & 0.624 & 0.673 & 0.822 & 0.564 & 0.704 & 0.405 & 0.603 \\
DetectGPT & 0.544 & 0.616 & 0.860 & \underline{0.913} & 0.607 & 0.693 & 0.781 & 0.404 & 0.371 & 0.670 & 0.530 & 0.612 \\
Binoculars & 0.885 & 0.874 & 0.826 & 0.667 & 0.802 & 0.813 & 0.757 & 0.886 & 0.827 & 0.775 & 0.493 & 0.720 \\
RADAR & 0.573 & 0.469 & 0.319 & 0.307 & 0.676 & 0.526 & 0.300 & 0.744 & 0.774 & 0.573 & 0.751 & 0.588 \\
BiScope & 0.953 & 0.910 & 0.833 & 0.840 & 0.825 & 0.804 & 0.735 & 0.945 & 0.853 & 0.846 & \underline{0.768} & \underline{0.832} \\
ImBD & \underline{0.973} & \underline{0.940} & 0.666 & 0.908 & \textbf{0.963} & \textbf{0.940} & \textbf{0.922} & \textbf{0.984} & \underline{0.943} & \underline{0.877} & 0.283 & 0.728 \\
Fast-DetectGPT & 0.881 & 0.881 & 0.837 & 0.713 & 0.809 & \underline{0.819} & 0.781 & 0.877 & 0.821 & 0.773 & 0.476 & 0.721 \\
AdaDetectGPT & 0.441 & 0.488 & \underline{0.880} & 0.888 & 0.499 & 0.510 & 0.660 & 0.419 & 0.339 & 0.627 & 0.598 & 0.585 \\
\textsc{Reader} & \textbf{0.977} & \textbf{0.951} & \textbf{0.961} & \textbf{0.975} & \underline{0.929} & \textbf{0.940} & \underline{0.919} & \underline{0.979} & \textbf{0.946} & \textbf{0.914} & \textbf{0.955} & \textbf{0.953} \\

\bottomrule
\end{tabular}
}
\end{sc}
\end{small}
\end{center}
\end{table*}

\begin{table*}[!t]
\vspace{-10pt}
\caption{\textsc{Read} benchmark: per-domain classification accuracy of baseline detectors and \textsc{Reader} under the per-domain configuration. For methods requiring a surrogate/reference LLM, we use Gemma-2-9B(-IT). \textbf{Bold} denotes the best performance in each column and \underline{underlined} denotes the second-best.}\label{tab:read_gemma_pre_domain}
\begin{center}
\begin{small}
\begin{sc}
\setlength{\tabcolsep}{2pt}
\renewcommand{\arraystretch}{1.05}
\resizebox{\textwidth}{!}{%
\begin{tabular}{l|cccccccccccc}
\toprule
\textbf{Method / Domain} & \textbf{CMV} & \textbf{ELI5} & \textbf{HellaSwag} & \textbf{ROC} & \textbf{SCI} & \textbf{SQuAD} & \textbf{TLDR} & \textbf{WP} & \textbf{XSum} & \textbf{Yelp} & \textbf{Rewrite} & \textbf{Avg.} \\
\midrule
RoBERTaBase & 0.844 & 0.798 & 0.796 & 0.809 & 0.725 & 0.777 & 0.754 & 0.850 & 0.827 & 0.771 & 0.864 & 0.818 \\
RoBERTaLarge & 0.817 & 0.787 & 0.767 & 0.749 & 0.741 & 0.750 & 0.729 & 0.807 & 0.807 & 0.721 & 0.865 & 0.798 \\
Likelihood & 0.878 & 0.845 & 0.860 & 0.872 & 0.756 & 0.789 & 0.798 & 0.875 & 0.748 & 0.771 & 0.866 & 0.836 \\
Entropy & 0.846 & 0.782 & 0.848 & 0.883 & 0.707 & 0.713 & 0.760 & 0.851 & 0.703 & 0.718 & 0.866 & 0.811 \\
LogRank & 0.881 & 0.843 & 0.861 & 0.870 & 0.756 & 0.786 & 0.798 & 0.873 & 0.751 & 0.770 & 0.865 & 0.836 \\
LRR & 0.797 & 0.764 & 0.780 & 0.793 & 0.652 & 0.701 & 0.722 & 0.831 & 0.725 & 0.696 & 0.864 & 0.785 \\
NPR & 0.763 & 0.724 & 0.915 & \underline{0.970} & 0.737 & 0.820 & 0.768 & 0.776 & 0.676 & 0.704 & 0.864 & 0.812 \\
DNAGPT & 0.811 & 0.797 & 0.665 & 0.676 & 0.737 & 0.664 & 0.684 & 0.835 & 0.699 & 0.736 & 0.864 & 0.770 \\
DetectGPT & 0.738 & 0.688 & \underline{0.918} & \underline{0.970} & 0.733 & 0.833 & 0.822 & 0.674 & 0.674 & 0.707 & 0.864 & 0.803 \\
Binoculars & 0.902 & 0.884 & 0.835 & 0.766 & 0.821 & 0.840 & 0.797 & 0.887 & 0.857 & 0.786 & 0.864 & 0.847 \\
RADAR & 0.721 & 0.665 & 0.830 & 0.882 & 0.727 & 0.628 & 0.727 & 0.786 & 0.774 & 0.621 & 0.864 & 0.779 \\
BiScope & 0.958 & 0.926 & 0.879 & 0.909 & 0.857 & 0.862 & 0.863 & 0.950 & 0.863 & 0.861 & \underline{0.869} & 0.888 \\
ImBD & \textbf{0.980} & \underline{0.947} & 0.880 & 0.937 & \textbf{0.969} & \textbf{0.942} & \textbf{0.933} & \textbf{0.995} & \textbf{0.956} & \underline{0.880} & 0.865 & \underline{0.920} \\
Fast-DetectGPT & 0.899 & 0.888 & 0.853 & 0.831 & 0.829 & 0.855 & 0.819 & 0.882 & 0.846 & 0.791 & 0.864 & 0.855 \\
AdaDetectGPT & 0.736 & 0.718 & 0.883 & 0.907 & 0.711 & 0.815 & 0.814 & 0.726 & 0.685 & 0.711 & 0.864 & 0.800 \\
\textsc{Reader} & \underline{0.977} & \textbf{0.951} & \textbf{0.961} & \textbf{0.975} & \underline{0.929} & \underline{0.940} & \underline{0.919} & \underline{0.979} & \underline{0.946} & \textbf{0.914} & \textbf{0.955} & \textbf{0.953} \\

\bottomrule
\end{tabular}
}
\end{sc}
\end{small}
\end{center}
\end{table*}

\subsection{Extended \textsc{Read} Benchmark Results}\label{app:extended-read-results}
Tables~\ref{tab:read_gemma}, \ref{tab:read_gemma_pre_domain}, \ref{tab:read_qwen}, and \ref{tab:read_qwen_pre_domain} report accuracies for all baselines and \textsc{Reader} across domains under four experimental settings, varying (i) the oracle-threshold strategy (global configuration or per-domain configuration) and (ii) the surrogate or reference LM used by LM-dependent detectors (Gemma-2-9B(-IT) or Qwen2.5-1.5B-Instruct). These tables provide the per-domain breakdown behind the in-distribution results summarized in Section~\ref{subsec:indis}.

\begin{table*}[!t]
\caption{\textsc{Read} benchmark: per-domain classification accuracy of baseline detectors and \textsc{Reader} under the global configuration. For methods requiring a surrogate/reference LLM, we use Qwen2.5-1.5B-Instruct. \textbf{Bold} denotes the best performance in each column and \underline{underlined} denotes the second-best.}\label{tab:read_qwen}
\begin{center}
\begin{small}
\begin{sc}
\setlength{\tabcolsep}{2pt}
\renewcommand{\arraystretch}{1.05}
\resizebox{\textwidth}{!}{%
\begin{tabular}{l|cccccccccccc}
\toprule
\textbf{Method / Domain} & \textbf{CMV} & \textbf{ELI5} & \textbf{HellaSwag} & \textbf{ROC} & \textbf{SCI} & \textbf{SQuAD} & \textbf{TLDR} & \textbf{WP} & \textbf{XSum} & \textbf{Yelp} & \textbf{Rewrite} & \textbf{Avg.} \\
\midrule
RoBERTaBase & 0.833 & 0.796 & 0.792 & 0.698 & 0.713 & 0.775 & 0.714 & 0.835 & 0.800 & 0.768 & 0.156 & 0.596 \\
RoBERTaLarge & 0.788 & 0.764 & 0.758 & 0.702 & 0.680 & 0.739 & 0.695 & 0.776 & 0.763 & 0.714 & 0.144 & 0.568 \\
Likelihood & 0.885 & 0.851 & 0.855 & 0.820 & 0.723 & 0.593 & 0.773 & 0.877 & 0.680 & 0.777 & 0.288 & 0.645 \\
Entropy & 0.838 & 0.789 & 0.825 & 0.641 & 0.640 & 0.559 & 0.748 & 0.848 & 0.623 & 0.713 & 0.340 & 0.616 \\
LogRank & 0.885 & 0.849 & 0.851 & 0.804 & 0.729 & 0.624 & 0.771 & 0.878 & 0.704 & 0.784 & 0.270 & 0.643 \\
LRR & 0.875 & 0.819 & 0.810 & 0.655 & 0.756 & 0.675 & 0.754 & 0.862 & 0.799 & 0.759 & 0.230 & 0.620 \\
NPR & 0.593 & 0.680 & 0.865 & \underline{0.909} & 0.593 & 0.541 & 0.752 & 0.455 & 0.450 & 0.734 & 0.487 & 0.611 \\
DNAGPT & 0.858 & 0.783 & 0.649 & 0.652 & 0.633 & 0.599 & 0.640 & 0.873 & 0.684 & 0.745 & 0.450 & 0.639 \\
DetectGPT & 0.585 & 0.637 & 0.857 & 0.901 & 0.605 & 0.572 & 0.810 & 0.435 & 0.422 & 0.711 & 0.491 & 0.607 \\
Binoculars & 0.846 & 0.845 & 0.715 & 0.709 & 0.762 & 0.699 & 0.656 & 0.822 & 0.786 & 0.766 & 0.235 & 0.610 \\
RADAR & 0.573 & 0.469 & 0.319 & 0.307 & 0.676 & 0.526 & 0.300 & 0.744 & 0.774 & 0.573 & \underline{0.751} & 0.588 \\
BiScope & 0.955 & \underline{0.917} & 0.808 & 0.853 & 0.904 & 0.786 & 0.808 & 0.961 & \underline{0.840} & \underline{0.870} & 0.747 & \underline{0.835} \\
ImBD & \underline{0.963} & 0.896 & 0.428 & 0.807 & \textbf{0.945} & \underline{0.813} & \underline{0.889} & \underline{0.972} & 0.824 & 0.820 & 0.302 & 0.677 \\
Fast-DetectGPT & 0.846 & 0.833 & 0.799 & 0.747 & 0.806 & 0.779 & 0.768 & 0.807 & 0.794 & 0.789 & 0.172 & 0.616 \\
AdaDetectGPT & 0.300 & 0.376 & \underline{0.871} & 0.877 & 0.387 & 0.419 & 0.448 & 0.341 & 0.331 & 0.448 & 0.693 & 0.550 \\
\textsc{Reader} & \textbf{0.977} & \textbf{0.951} & \textbf{0.961} & \textbf{0.975} & \underline{0.929} & \textbf{0.940} & \textbf{0.919} & \textbf{0.979} & \textbf{0.946} & \textbf{0.914} & \textbf{0.955} & \textbf{0.953} \\

\bottomrule

\end{tabular}
}
\end{sc}
\end{small}
\end{center}
\end{table*}

\begin{table*}[!t]
\caption{\textsc{Read} benchmark: per-domain classification accuracy of baseline detectors and \textsc{Reader} under the per-domain configuration. For methods requiring a surrogate/reference LLM, we use Qwen2.5-1.5B-Instruct. \textbf{Bold} denotes the best performance in each column and \underline{underlined} denotes the second-best.}\label{tab:read_qwen_pre_domain}
\begin{center}
\begin{small}
\begin{sc}
\setlength{\tabcolsep}{2pt}
\renewcommand{\arraystretch}{1.05}
\resizebox{\textwidth}{!}{%
\begin{tabular}{l|cccccccccccc}
\toprule
\textbf{Method / Domain} & \textbf{CMV} & \textbf{ELI5} & \textbf{HellaSwag} & \textbf{ROC} & \textbf{SCI} & \textbf{SQuAD} & \textbf{TLDR} & \textbf{WP} & \textbf{XSum} & \textbf{Yelp} & \textbf{Rewrite} & \textbf{Avg.} \\
\midrule
RoBERTaBase & 0.844 & 0.798 & 0.796 & 0.809 & 0.725 & 0.777 & 0.754 & 0.850 & 0.827 & 0.771 & 0.864 & 0.818 \\
RoBERTaLarge & 0.817 & 0.787 & 0.767 & 0.749 & 0.741 & 0.750 & 0.729 & 0.807 & 0.807 & 0.721 & 0.865 & 0.798 \\
Likelihood & 0.889 & 0.856 & 0.857 & 0.850 & 0.778 & 0.759 & 0.821 & 0.880 & 0.767 & 0.793 & 0.864 & 0.839 \\
Entropy & 0.838 & 0.792 & 0.830 & 0.769 & 0.717 & 0.679 & 0.765 & 0.858 & 0.709 & 0.741 & 0.864 & 0.801 \\
LogRank & 0.886 & 0.852 & 0.855 & 0.850 & 0.778 & 0.766 & 0.821 & 0.880 & 0.785 & 0.789 & 0.864 & 0.840 \\
LRR & 0.880 & 0.834 & 0.837 & 0.766 & 0.768 & 0.726 & 0.768 & 0.863 & 0.808 & 0.764 & 0.864 & 0.823 \\
NPR & 0.771 & 0.728 & 0.913 & \underline{0.955} & 0.752 & 0.771 & 0.794 & 0.774 & 0.680 & 0.738 & 0.864 & 0.813 \\
DNAGPT & 0.872 & 0.833 & 0.739 & 0.739 & 0.766 & 0.710 & 0.716 & 0.878 & 0.743 & 0.757 & 0.864 & 0.803 \\
DetectGPT & 0.733 & 0.696 & \underline{0.916} & 0.953 & 0.729 & 0.808 & 0.837 & 0.659 & 0.672 & 0.718 & 0.864 & 0.800 \\
Binoculars & 0.856 & 0.849 & 0.747 & 0.717 & 0.815 & 0.760 & 0.760 & 0.825 & 0.818 & 0.782 & 0.864 & 0.813 \\
RADAR & 0.721 & 0.665 & 0.830 & 0.882 & 0.727 & 0.628 & 0.727 & 0.786 & 0.774 & 0.621 & 0.864 & 0.779 \\
BiScope & 0.959 & \underline{0.926} & 0.893 & 0.914 & 0.908 & 0.826 & 0.865 & 0.961 & 0.851 & \underline{0.873} & \underline{0.870} & \underline{0.892} \\
ImBD & \underline{0.966} & 0.900 & 0.812 & 0.866 & \textbf{0.953} & \underline{0.884} & \underline{0.897} & \textbf{0.987} & \underline{0.883} & 0.830 & 0.865 & 0.888 \\
Fast-DetectGPT & 0.847 & 0.840 & 0.808 & 0.766 & 0.813 & 0.800 & 0.789 & 0.815 & 0.809 & 0.798 & 0.864 & 0.825 \\
AdaDetectGPT & 0.722 & 0.661 & 0.893 & 0.893 & 0.713 & 0.828 & 0.808 & 0.747 & 0.722 & 0.661 & 0.864 & 0.796 \\
\textsc{Reader} & \textbf{0.977} & \textbf{0.951} & \textbf{0.961} & \textbf{0.975} & \underline{0.929} & \textbf{0.940} & \textbf{0.919} & \underline{0.979} & \textbf{0.946} & \textbf{0.914} & \textbf{0.955} & \textbf{0.953} \\

\bottomrule
\end{tabular}
}
\end{sc}
\end{small}
\end{center}
\end{table*}

For any threshold-based detector that outputs a scalar score $s(x)$, the oracle threshold on an evaluation set $\mathcal{S}$ is defined as
\[
\begin{aligned}
\tau^{*}(\mathcal{S})
&= \arg\max_{\tau}\; \frac{1}{|\mathcal{S}|}
\sum_{(x_i,y_i)\in\mathcal{S}}
\mathbb{I}\!\left[\hat{y}_i(\tau)=y_i\right],\\
\hat{y}_i(\tau)
&= \mathbb{I}\!\left[s(x_i)\ge \tau\right].
\end{aligned}
\]
i.e., we select the threshold that maximizes accuracy on $\mathcal{S}$. In the global configuration, $\mathcal{S}$ is the full test set; in the per-domain configuration, $\mathcal{S}$ is restricted to each domain separately. For Fast-DetectGPT and AdaDetectGPT, which require a \emph{sampling} LM and a \emph{scoring} LM for detection, we use \textit{Gemma-2-9B-IT} (sampling) and \textit{Gemma-2-9B} (scoring) under the Gemma setting following \citet{zhou2025adadetect}, and \textit{Qwen2.5-1.5B-Instruct} as both sampling and scoring LMs under the Qwen setting.

A notable exception arises on \textsc{Rewrite}, where several threshold-based methods exhibit a significant drop when switching oracle settings. This behavior is largely explained by the evaluation design of \textsc{Rewrite}, which serves as a challenging label-imbalanced setting with a large majority of samples generated by AI. Under such skew, some baselines tend to trivially predict all samples as AI when optimizing thresholds, yielding a ``cheated'' high accuracy under per-domain oracle thresholding but failing when a single global threshold is imposed. In contrast, \textsc{Reader} remains robust on \textsc{Rewrite}, since its decisions are produced directly from model reasoning rather than manipulatable threshold calibration.
\paragraph{FPR/TPR results.}
Because the \textsc{Read} test set is balanced, accuracy is the main metric in the body of the paper. We additionally report false positive rate (FPR; human text incorrectly flagged as AI) and true positive rate (TPR; AI text correctly detected) here. Tables~\ref{tab:fpr_tpr_read_global} and \ref{tab:fpr_tpr_read_domain} show that \textsc{Reader} combines the lowest or near-lowest FPR with the strongest TPR, whereas several threshold-based baselines obtain competitive FPR only by missing many AI-generated samples.

\begin{table*}[!b]
\vspace{-0.5em}
\begin{center}
\captionsetup{hypcap=false}
\begin{minipage}[!t]{0.48\textwidth}
    \captionof{table}{False positive rate (FPR) and true positive rate (TPR) of various detectors on the \textsc{Read} test set (global oracle threshold configuration for baselines). For methods requiring a surrogate LLM, we report results using both Gemma-2-9B(-IT) and Qwen2.5-1.5B-Instruct. The best performance in each column is \textbf{bolded}, and the second-best is \underline{underlined}.}
    \label{tab:fpr_tpr_read_global}
    \begin{center}
        \begin{small}
            \setlength{\tabcolsep}{2.5pt}
            \begin{tabular}{lcccc}
            \toprule
            \textbf{Method} & \multicolumn{2}{c}{\textbf{FPR} $\downarrow$} & \multicolumn{2}{c}{\textbf{TPR} $\uparrow$} \\
            \cmidrule(lr){2-3}\cmidrule(lr){4-5}
            & \textbf{Gemma} & \textbf{Qwen} & \textbf{Gemma} & \textbf{Qwen} \\
            \midrule
            \quad \textsc{Likelihood}        & 0.126 & 0.134 & 0.411 & 0.425 \\
            \quad \textsc{Entropy}           & 0.198 & 0.180 & 0.440 & 0.412 \\
            \quad \textsc{LogRank}           & 0.124 & 0.125 & 0.403 & 0.411 \\
            \quad \textsc{LRR}               & 0.232 & 0.132 & 0.500 & 0.373 \\
            \quad \textsc{NPR}               & 0.304 & 0.361 & 0.541 & 0.583 \\
            \quad \textsc{DNAGPT}            & 0.281 & 0.281 & 0.486 & 0.559 \\
            \quad \textsc{DetectGPT}         & 0.367 & 0.360 & 0.590 & 0.575 \\
            \quad \textsc{Binoculars}        & 0.115 & 0.149 & 0.554 & 0.369 \\
            \quad \textsc{BiScope}           & 0.166 & 0.152 & \underline{0.830} & \underline{0.823} \\
            \quad \textsc{ImBD}              & 0.105 & 0.211 & 0.560 & 0.564 \\
            \quad \textsc{Fast-DetectGPT}    & 0.098 & \underline{0.049} & 0.540 & 0.280 \\
            \quad \textsc{AdaDetectGPT}      & 0.553 & 0.722 & 0.724 & 0.822 \\
            \midrule
            \quad \textsc{RoBERTa-base}      & \multicolumn{2}{c}{0.057} & \multicolumn{2}{c}{0.248} \\
            \quad \textsc{RoBERTa-large}     & \multicolumn{2}{c}{\textbf{0.045}} & \multicolumn{2}{c}{0.181} \\
            \quad \textsc{RADAR}             & \multicolumn{2}{c}{0.651} & \multicolumn{2}{c}{\underline{0.828}} \\
            \midrule
            \quad \textsc{Reader}   & \multicolumn{2}{c}{\underline{0.050}} & \multicolumn{2}{c}{\textbf{0.955}} \\
            \bottomrule
        \end{tabular}
    \end{small}
  \end{center}
  \vskip -0.1in
\end{minipage}\hfill
\begin{minipage}[!t]{0.48\textwidth}

    \captionof{table}{False positive rate (FPR) and true positive rate (TPR) of various detectors on the \textsc{Read} test set (per-domain oracle threshold configuration for baselines). For methods requiring a surrogate LLM, we report results using both Gemma-2-9B(-IT) and Qwen2.5-1.5B-Instruct. The best performance in each column is \textbf{bolded}, and the second-best is \underline{underlined}.}
    \label{tab:fpr_tpr_read_domain}
    \begin{center}
        \begin{small}
            \setlength{\tabcolsep}{2.5pt}
            \begin{tabular}{lcccc}
            \toprule
            \textbf{Method} & \multicolumn{2}{c}{\textbf{FPR} $\downarrow$} & \multicolumn{2}{c}{\textbf{TPR} $\uparrow$} \\
            \cmidrule(lr){2-3}\cmidrule(lr){4-5}
            & \textbf{Gemma} & \textbf{Qwen} & \textbf{Gemma} & \textbf{Qwen} \\
            \midrule
            \quad \textsc{Likelihood}        & 0.112 & 0.103 & 0.775 & 0.774 \\
            \quad \textsc{Entropy}           & 0.101 & 0.123 & 0.708 & 0.711 \\
            \quad \textsc{LogRank}           & 0.113 & 0.103 & 0.776 & 0.775 \\
            \quad \textsc{LRR}               & 0.147 & 0.126 & 0.702 & 0.759 \\
            \quad \textsc{NPR}               & 0.094 & 0.096 & 0.701 & 0.706 \\
            \quad \textsc{DNAGPT}            & 0.145 & 0.158 & 0.671 & 0.753 \\
            \quad \textsc{DetectGPT}         & 0.113 & 0.107 & 0.700 & 0.689 \\
            \quad \textsc{Binoculars}        & 0.094 & 0.118 & 0.781 & 0.729 \\
            \quad \textsc{BiScope}           & 0.089 & 0.095 & {0.872} & {0.883} \\
            \quad \textsc{ImBD}              & 0.101 & 0.130 & \underline{0.949} & \underline{0.904} \\
            \quad \textsc{Fast-DetectGPT}    & \underline{0.079} & \underline{0.094} & 0.783 & 0.728 \\
            \quad \textsc{AdaDetectGPT}      & 0.108 & 0.115 & 0.689 & 0.691 \\
            \midrule
            \quad \textsc{RoBERTa-base}      & \multicolumn{2}{c}{0.098} & \multicolumn{2}{c}{0.720} \\
            \quad \textsc{RoBERTa-large}     & \multicolumn{2}{c}{0.106} & \multicolumn{2}{c}{0.690} \\
            \quad \textsc{RADAR}             & \multicolumn{2}{c}{0.149} & \multicolumn{2}{c}{0.688} \\
            \midrule
            \quad \textsc{Reader}   & \multicolumn{2}{c}{\textbf{0.050}} & \multicolumn{2}{c}{\textbf{0.955}} \\
            \bottomrule
        \end{tabular}
    \end{small}
  \end{center}
  \vskip -0.1in
\end{minipage}
\end{center}
\vspace{-0.8em}
\end{table*}

\subsection{Additional OOD Evaluation Details}\label{app:ood-details}
This subsection expands the OOD evaluation described in Section~\ref{subsec:outofdis}. We evaluate two settings: (i) unseen generators and (ii) cross-lingual and adversarial robustness.

\paragraph{Unseen generators test.} First, the unseen generators test follows the evaluation protocol of \citet{bao2024fastdetectgpt}. We sample 150 human-written paragraphs from each dataset and generate LLM-authored continuations conditioned on either 120-token or 200-token prefixes. The machine-generated texts are produced by LLMs that are \emph{unseen} during training, including \ul{\textit{Claude-Haiku-3.5}}, \ul{\textit{Gemini-2.5-Flash}}, \ul{\textit{GPT-4o}}, and \ul{\textit{GPT-4}}. The corresponding human texts are drawn from four diverse domains: \ul{\textit{XSum}}, \ul{\textit{WritingPrompts}}, \ul{\textit{Yelp}}, and \ul{\textit{Essay}} \citep{verma2024ghostbuster}. In particular, the \textit{Essay} domain, which consists of high-school and university-level writing, is entirely absent from the training data and therefore provides a domain-level OOD evaluation. Tables~\ref{tab:blackbox_gemma} and \ref{tab:blackbox_qwen} report the corresponding Gemma- and Qwen-based surrogate results.

\begin{table*}[!tbp]
\caption{Out-of-distribution detection accuracy. Baselines are evaluated under the per-domain configuration, with Gemma employed as the surrogate LLM. The best performance in each column is \textbf{bolded}, and the second-best is \underline{underlined}.}
\label{tab:blackbox_gemma}
\begin{center}
\begin{small}
\begin{sc}
\setlength{\tabcolsep}{2.5pt}
\renewcommand{\arraystretch}{1.05}
\resizebox{\textwidth}{!}{
\begin{tabular}{l|cccc:c|cccc:c|cccc:c}
\toprule
\multicolumn{1}{c}{\textbf{Source LLM}} &
\multicolumn{5}{c}{\textbf{Gemini-2.5-Flash}} &
\multicolumn{5}{c}{\textbf{Claude-3.5-Haiku}} &
\multicolumn{5}{c}{\textbf{GPT-4o}} \\
\cmidrule(lr){1-1}\cmidrule(lr){2-5}\cmidrule(lr){6-6}\cmidrule(lr){7-10}\cmidrule(lr){11-11}\cmidrule(lr){12-15}\cmidrule(lr){16-16}
\textbf{Method / Domain} & \textbf{XSum} & \textbf{Writing} & \textbf{Yelp} & \textbf{Essay} & \textbf{Avg.}
& \textbf{XSum} & \textbf{Writing} & \textbf{Yelp} & \textbf{Essay} & \textbf{Avg.}
& \textbf{XSum} & \textbf{Writing} & \textbf{Yelp} & \textbf{Essay} & \textbf{Avg.} \\
\midrule
RoBERTaBase     & 0.580 & 0.547 & 0.573 & 0.740 & 0.610 & 0.550 & 0.603 & 0.560 & 0.587 & 0.575 & 0.543 & 0.547 & 0.587 & 0.587 & 0.566 \\
RoBERTaLarge     & 0.650 & 0.640 & 0.613 & 0.743 & 0.662 & 0.567 & 0.593 & 0.543 & 0.600 & 0.576 & 0.523 & 0.583 & 0.527 & 0.633 & 0.567 \\
Likelihood       & 0.607 & 0.570 & 0.663 & 0.713 & 0.638 & 0.537 & 0.637 & 0.700 & 0.660 & 0.633 & 0.540 & 0.713 & 0.767 & 0.723 & 0.686 \\
Entropy          & 0.570 & 0.567 & 0.597 & 0.657 & 0.598 & 0.593 & 0.590 & 0.577 & 0.630 & 0.598 & 0.547 & 0.653 & 0.680 & 0.673 & 0.638 \\
LogRank          & 0.600 & 0.563 & 0.663 & 0.690 & 0.629 & 0.530 & 0.627 & 0.683 & 0.657 & 0.624 & 0.530 & 0.687 & 0.760 & 0.720 & 0.674 \\
LRR              & 0.573 & 0.570 & 0.557 & 0.637 & 0.584 & 0.537 & 0.567 & 0.557 & 0.623 & 0.571 & 0.540 & 0.597 & 0.650 & 0.643 & 0.607 \\
NPR              & 0.550 & 0.687 & 0.550 & 0.927 & 0.678 & 0.613 & 0.730 & 0.580 & 0.803 & 0.682 & 0.633 & 0.763 & 0.633 & 0.837 & 0.717 \\
DNAGPT           & 0.653 & 0.560 & 0.700 & 0.853 & 0.692 & 0.573 & 0.640 & 0.773 & 0.807 & 0.698 & 0.547 & 0.767 & 0.873 & 0.867 & 0.763 \\
DetectGPT        & 0.567 & 0.687 & 0.630 & 0.900 & 0.696 & 0.663 & 0.740 & 0.703 & 0.807 & 0.728 & 0.677 & 0.733 & 0.670 & 0.833 & 0.728 \\
Binoculars       & \underline{0.800} & 0.870 & \textbf{0.913} & 0.957 & 0.885 & 0.833 & 0.880 & 0.920 & \underline{0.903} & \underline{0.884} & 0.833 & 0.900 & 0.940 & \underline{0.943} & 0.904 \\
RADAR            & 0.760 & 0.540 & 0.603 & 0.583 & 0.622 & \underline{0.857} & 0.720 & 0.777 & 0.847 & 0.800 & \underline{0.923} & 0.767 & 0.793 & 0.780 & 0.816 \\
BiScope & 0.757 & 0.763 & 0.713 & 0.863 & 0.774 & 0.773 & 0.760 & 0.737 & 0.770 & 0.760 & 0.893 & 0.883 & 0.860 & 0.820 & 0.864 \\
BiScope$^{*}$    & 0.680 & 0.763 & 0.807 & 0.843 & 0.773 & 0.780 & 0.787 & 0.707 & 0.817 & 0.772 & 0.860 & 0.890 & 0.910 & 0.900 & 0.890 \\
ImBD & 0.573 & 0.647 & 0.683 & 0.710 & 0.653 & 0.720 & 0.743 & 0.750 & 0.787 & 0.750 & 0.730 & 0.860 & 0.820 & 0.693 & 0.776 \\
ImBD$^{*}$       & 0.753 & \textbf{0.910} & 0.907 & \textbf{0.980} & \underline{0.887} & 0.823 & \underline{0.917} & 0.913 & 0.873 & 0.882 & 0.870 & \underline{0.953} & \textbf{0.960} & 0.933 & \underline{0.929} \\
Fast-DetectGPT   & 0.787 & 0.863 & \underline{0.910} & 0.960 & 0.880 & 0.833 & 0.873 & \textbf{0.927} & 0.897 & 0.882 & 0.833 & 0.903 & 0.943 & \underline{0.943} & 0.906 \\
AdaDetectGPT     & 0.553 & 0.663 & 0.583 & 0.600 & 0.600 & 0.530 & 0.600 & 0.537 & 0.597 & 0.566 & 0.550 & 0.617 & 0.603 & 0.573 & 0.586 \\
AdaDetectGPT$^{*}$ & 0.783 & \underline{0.887} & \textbf{0.913} & \underline{0.967} & \underline{0.887} & 0.837 & 0.870 & \underline{0.923} & 0.900 & 0.882 & 0.827 & 0.897 & 0.940 & 0.940 & 0.901 \\
\textbf{Reader}           & \textbf{0.960} & 0.857 & 0.903 & \textbf{0.980} & \textbf{0.925} & \textbf{0.977} & \textbf{0.923} & 0.873 & \textbf{0.963} & \textbf{0.934} & \textbf{0.987} & \textbf{0.980} & \underline{0.957} & \textbf{0.973} & \textbf{0.974} \\
\midrule
\end{tabular}
}
\end{sc}
\end{small}
\end{center}
\vskip -0.1in
\end{table*}

\begin{table*}[!tbp]
\caption{Out-of-distribution detection accuracy. Baselines are evaluated under the per-domain configuration, with Qwen2.5-1.5B-Instruct employed as the surrogate LLM. The best performance in each column is \textbf{bolded}, and the second-best is \underline{underlined}.}
\label{tab:blackbox_qwen}
\begin{center}
\begin{small}
\begin{sc}
\setlength{\tabcolsep}{2.5pt}
\renewcommand{\arraystretch}{1.05}
\resizebox{\textwidth}{!}{%
\begin{tabular}{l|cccc:c|cccc:c|cccc:c}
\toprule
\multicolumn{1}{c|}{\textbf{Source LLM}} & \multicolumn{5}{c|}{\textbf{Gemini-2.5-Flash}} & \multicolumn{5}{c|}{\textbf{Claude-3.5-Haiku}} & \multicolumn{5}{c}{\textbf{GPT-4o}} \\
\cmidrule(lr){1-1}\cmidrule(lr){2-5}\cmidrule(lr){6-6}\cmidrule(lr){7-10}\cmidrule(lr){11-11}\cmidrule(lr){12-15}\cmidrule(lr){16-16}
{\textbf{Method / Domain}} & \textbf{XSum} & \textbf{Writing} & \textbf{Yelp} & \textbf{Essay} & \textbf{Avg.} & \textbf{XSum} & \textbf{Writing} & \textbf{Yelp} & \textbf{Essay} & \textbf{Avg.} & \textbf{XSum} & \textbf{Writing} & \textbf{Yelp} & \textbf{Essay} & \textbf{Avg.} \\
\midrule
RoBERTa-base & 0.580 & 0.547 & 0.573 & 0.740 & 0.610 & 0.550 & 0.603 & 0.560 & 0.587 & 0.575 & 0.543 & 0.547 & 0.587 & 0.587 & 0.566 \\
RoBERTa-large & 0.650 & 0.640 & 0.613 & 0.743 & 0.662 & 0.567 & 0.593 & 0.543 & 0.600 & 0.576 & 0.523 & 0.583 & 0.527 & 0.633 & 0.567 \\
Likelihood & 0.573 & 0.570 & 0.640 & 0.663 & 0.612 & 0.560 & 0.640 & 0.670 & 0.653 & 0.631 & 0.570 & 0.733 & 0.767 & 0.740 & 0.702 \\
Entropy & 0.530 & 0.560 & 0.603 & 0.633 & 0.582 & 0.573 & 0.580 & 0.613 & 0.660 & 0.607 & 0.583 & 0.650 & 0.693 & 0.693 & 0.655 \\
LogRank & 0.570 & 0.567 & 0.637 & 0.653 & 0.607 & 0.553 & 0.637 & 0.663 & 0.663 & 0.629 & 0.587 & 0.710 & 0.753 & 0.737 & 0.697 \\
LRR & 0.570 & 0.547 & 0.600 & 0.617 & 0.583 & 0.580 & 0.597 & 0.637 & 0.657 & 0.618 & 0.577 & 0.627 & 0.700 & 0.673 & 0.644 \\
NPR & 0.580 & 0.670 & 0.543 & 0.890 & 0.671 & 0.703 & 0.723 & 0.580 & 0.800 & 0.702 & 0.703 & 0.800 & 0.643 & 0.843 & 0.747 \\
DNAGPT & 0.583 & 0.573 & 0.637 & 0.763 & 0.639 & 0.627 & 0.640 & 0.713 & 0.743 & 0.681 & 0.590 & 0.777 & 0.833 & 0.833 & 0.758 \\
DetectGPT & 0.600 & 0.660 & 0.527 & 0.897 & 0.671 & 0.720 & 0.733 & 0.603 & 0.833 & 0.723 & 0.713 & 0.780 & 0.663 & 0.850 & 0.752 \\
Binoculars & 0.613 & 0.617 & 0.637 & 0.617 & 0.621 & 0.543 & 0.613 & 0.693 & 0.523 & 0.593 & 0.553 & 0.683 & 0.753 & 0.690 & 0.670 \\
RADAR & 0.760 & 0.540 & 0.603 & 0.583 & 0.622 & 0.857 & 0.720 & 0.777 & 0.847 & 0.800 & 0.923 & 0.767 & 0.793 & 0.780 & 0.816 \\
BiScope & \underline{0.850} & 0.823 & 0.740 & \underline{0.977} & 0.847 & 0.850 & 0.800 & 0.763 & 0.857 & 0.817 & 0.927 & 0.923 & \underline{0.880} & 0.933 & 0.916 \\
BiScope$^{*}$ & 0.827 & \textbf{0.867} & \underline{0.863} & 0.973 & \underline{0.883} & \underline{0.890} & \underline{0.877} & \textbf{0.893} & \underline{0.953} & \underline{0.903} & \underline{0.930} & \underline{0.950} & \textbf{0.957} & \textbf{0.983} & \underline{0.955} \\
ImBD & 0.537 & 0.603 & 0.583 & 0.613 & 0.584 & 0.657 & 0.697 & 0.580 & 0.587 & 0.630 & 0.667 & 0.803 & 0.727 & 0.693 & 0.723 \\
ImBD$^{*}$ & 0.693 & 0.767 & 0.747 & 0.903 & 0.777 & 0.750 & 0.760 & 0.830 & 0.887 & 0.807 & 0.767 & 0.890 & 0.853 & 0.913 & 0.856 \\
Fast-DetectGPT & 0.633 & 0.607 & 0.630 & 0.617 & 0.622 & 0.547 & 0.617 & 0.700 & 0.533 & 0.599 & 0.550 & 0.683 & 0.757 & 0.697 & 0.672 \\
AdaDetectGPT & 0.557 & 0.547 & 0.670 & 0.707 & 0.620 & 0.657 & 0.670 & 0.713 & 0.707 & 0.687 & 0.640 & 0.687 & 0.757 & 0.690 & 0.693 \\
AdaDetectGPT$^{*}$ & 0.633 & 0.613 & 0.623 & 0.597 & 0.617 & 0.550 & 0.627 & 0.693 & 0.573 & 0.611 & 0.537 & 0.683 & 0.743 & 0.723 & 0.672 \\
\textsc{Reader} & \textbf{0.960} & \underline{0.857} & \textbf{0.903} & \textbf{0.980} & \textbf{0.925} & \textbf{0.977} & \textbf{0.923} & \underline{0.873} & \textbf{0.963} & \textbf{0.934} & \textbf{0.987} & \textbf{0.980} & \textbf{0.957} & \underline{0.973} & \textbf{0.974} \\
\bottomrule
\end{tabular}
}
\end{sc}
\end{small}
\end{center}
\end{table*}

Supervised detectors such as ImBD and AdaDetectGPT suffer considerably from distribution shift: their accuracy deteriorates sharply when the test data differs from the training data \textsc{Read} in either domain or source model. For this reason, Section~\ref{subsec:outofdis} reports the stronger target-adapted setting for those supervised baselines, while Tables~\ref{tab:blackbox_gemma} and \ref{tab:blackbox_qwen} also show the fixed-training variant trained only on \textsc{Read}. \textsc{Reader} considerably outperforms both zero-shot methods and supervised detectors trained on \textsc{Read}. Specifically, it outperforms these detectors by \textbf{5}-\textbf{35} points in the average detection accuracy across domains.

\paragraph{Cross-lingual and adversarial robustness test.} The other OOD setting evaluates cross-lingual and adversarial robustness based on the NLPCC benchmark. Unlike the previous OOD settings, which focus on unseen generators and domains, this setting introduces a language-level distribution shift: the test texts are in Chinese, whereas Chinese data is entirely absent from our training set. In addition, we consider three adversarial settings. In the mixed-content attack, LLM-generated documents are mixed with human-written content to confuse the detector at the paragraph or document level. In the paraphrase attack, LLM-generated Chinese texts are translated into English and then translated back into Chinese, simulating machine-translation-assisted paraphrasing or rewriting by non-native speakers. In the perturbation attack, characters in the LLM-generated Chinese texts are replaced with visually similar characters, simulating noisy human edits, typographical errors, or intentional character-level obfuscation. This benchmark therefore evaluates a stronger form of robustness, where \textsc{Reader} must handle both an unseen language and deliberate adversarial input manipulation.

\begin{table}[t]
\centering
\caption{Performance of \textsc{Reader} on cross-lingual and adversarial robustness benchmark.}
\label{tab:nlpcc_cross_lingual_adversarial}
\begin{tabular}{lc}
\toprule
\textbf{Setting} & \textbf{Accuracy} \\
\midrule
Normal & 93.4\% \\
Mixed attack & 92.1\% \\
Paraphrase attack & 91.8\% \\
Perturbation attack & 92.9\% \\
\bottomrule
\end{tabular}
\end{table}

Table~\ref{tab:nlpcc_cross_lingual_adversarial} shows that \textsc{Reader} retains over 91\% accuracy across all three adversarial variants, with only a small drop relative to the normal Chinese setting. This supports the OOD conclusion in Section~\ref{subsec:outofdis}: the detector is not relying solely on the English-domain artifacts seen during training.
\begin{table}[t]
\centering
\caption{Detection accuracy of our base model (Qwen2.5-1.5B-Instruct), SFT model (\textsc{Reader}-SFT) and \textsc{Reader}.}
\label{tab:ablation_read}
  \begin{center}
    \begin{small}
      \begin{sc}
      \setlength{\tabcolsep}{4pt}
\begin{tabular}{l c c c c}
\toprule
 & \textbf{Non-CoT} & & \textbf{CoT}  \\
\cmidrule(lr){2-2} \cmidrule(lr){4-4}
\textbf{Method} & \textbf{Avg.} & & \textbf{Avg.} & \\
\midrule
Qwen2.5-1.5B-Instruct & 0.495 & & 0.503 {\tiny(0.248)}  \\
$\hookrightarrow$ + SFT (\textsc{Reader}-SFT) & 0.710 & &  0.812  \\
$\hookrightarrow$ \textbf{+ GRPO (\textsc{Reader})} & \textbf{0.943} & & \textbf{0.953}   \\
\bottomrule
\end{tabular}
\end{sc}
\end{small}
\end{center}
\end{table}
\subsection{Ablation Details}\label{sec:ablation}\label{app:ablation-details}
Tables~\ref{tab:ablation_read} and \ref{tab:ablation_models} provide the full ablation results supporting the reasoning-enhanced design of \textsc{Reader}. The comparison isolates the two training stages and the inference-time prompting strategy: the base model is Qwen2.5-1.5B-Instruct, \textsc{Reader}-SFT is trained with one-step SFT, and \textsc{Reader} is trained with both SFT and GRPO. Each model is evaluated under Non-CoT prompting, which requires only the final verdict (Prompt~\ref{app:prompt-noncot}), and CoT prompting, which explicitly instructs the model to reason before reaching a final decision (Prompt~\ref{app:prompt-balanced}). The \textsc{Read} benchmark results are reported in Table~\ref{tab:ablation_read}; the OOD results are reported in Table~\ref{tab:ablation_models}.

The ablation results show four patterns. First, the base Qwen model performs poorly: under Non-CoT prompting, it classifies almost all samples as AI-generated; under CoT prompting, it frequently fails to follow the required output format, resulting in unparseable responses. To make this failure mode explicit, we report both the accuracy over valid, extractable predictions and the standard accuracy across the entire test set, with the latter shown in parentheses. Both metrics remain low and are comparable to random guessing. Second, \textsc{Reader}-SFT reliably generates valid responses and achieves substantially higher accuracy, addressing both limitations of the base model. Third, GRPO further boosts performance: \textsc{Reader} improves by roughly $15$--$20$ percentage points over \textsc{Reader}-SFT and achieves over $90\%$ accuracy across all tasks, highlighting the benefit of GRPO for strengthening the model's reasoning behavior. Fourth, CoT-style prompting consistently improves detection performance; in particular, the CoT setting improves \textsc{Reader}-SFT by about 10 percentage points over its Non-CoT counterpart. These results show that the proposed training and inference pipeline is effective for both producing valid rationale-label outputs and improving detection accuracy. 
\begin{table*}[t]
\centering
\caption{Detection accuracy of our base model (Qwen2.5-1.5B-Instruct), SFT model (\textsc{Reader}-SFT) and \textsc{Reader} under both CoT and non-CoT prompting strategies.}
\label{tab:ablation_models}
  \begin{center}
    \begin{small}
      \begin{sc}
\setlength{\tabcolsep}{0.5pt}

\begin{tabular}{l cccc c cccc}
\toprule
 & \multicolumn{4}{c}{\textbf{Non-CoT}} & & \multicolumn{4}{c}{\textbf{CoT}} \\
\cmidrule(lr){2-5} \cmidrule(lr){7-10}

\textbf{Method} & \textbf{Gemini-2.5} & \textbf{Claude-3.5} & \textbf{GPT-4o} & \textbf{Avg.} &
       & \textbf{Gemini-2.5} & \textbf{Claude-3.5} & \textbf{GPT-4o} & \textbf{Avg.} \\
\midrule

Qwen2.5-1.5B-Instruct
    & 0.487 & 0.501 & 0.480 & 0.489
    &
    & 0.440 {\tiny (0.193)} & 0.464 {\tiny(0.201)} & 0.403 {\tiny(0.188)} & 0.435 {\tiny(0.194)} \\

$\hookrightarrow$ + SFT (\textsc{Reader}-SFT)
    & 0.628 & 0.657 & 0.695 & 0.660
    &
    & 0.709 & 0.718 & 0.778 & 0.735 \\

$\hookrightarrow$ \textbf{+ GRPO (\textsc{Reader})}
    & \textbf{0.895} & \textbf{0.922} & \textbf{0.958} & \textbf{0.925}
    &
    & \textbf{0.925} & \textbf{0.934} & \textbf{0.974} & \textbf{0.944} \\

\bottomrule
\end{tabular}
      \end{sc}
    \end{small}
  \end{center}
  \vskip -0.1in

\end{table*} 
\begin{table}[t]
\centering
\small
\caption{Output-level consistency between \textsc{Reader}'s generated rationale and final verdict.}
\label{tab:rationale_consistency}
\begin{tabular}{lc}
\toprule
Metric & Value \\
\midrule
Rationale--verdict match rate & 1.000 \\
No explicit label-supporting statement & 0 \\
\bottomrule
\end{tabular}
\end{table}
\subsection{Diagnostics for Rationale--Verdict Coupling}
\label{app:rationale_coupling}
In this section, we conduct diagnostic analyses that test whether the rationales are aligned with, predictive of, and informative about \textsc{Reader}'s final predictions. 

    
\paragraph{Output-level consistency.}
We first examine whether the generated rationale contains an explicit label-supporting statement that matches the final verdict. This analysis checks whether the rationale is output-consistent with the prediction, and whether the model produces label-detached rationales. As shown in Table~\ref{tab:rationale_consistency}, the rationale-level label matches the final verdict in all evaluated examples, and no example lacks an explicit label-supporting statement. This verifies that \textsc{Reader}'s rationales are consistently aligned with the final verdict at the output level.

\begin{table*}[t]
\centering
\small
\setlength{\tabcolsep}{20pt}
\caption{Predicting \textsc{Reader}'s final verdict from its generated rationale using external BERT embeddings and logistic regression. Masked rationales remove explicit AI/HUMAN label words before encoding.}
\label{tab:rationale_predictiveness}
\begin{tabular}{lccc}
\toprule
Setting & Accuracy & AUROC & F1 \\
\midrule
Original rationales & 0.9997 & 1.0000 & 0.9997 \\
Masked rationales & 0.9982 & 1.0000 & 0.9982 \\
Original rationales on correctly predicted examples & 0.9997 & 1.0000 & 0.9997 \\
Original rationales on wrongly predicted examples & 0.9890 & 0.9998 & 0.9896 \\
Masked rationales on correctly predicted examples & 0.9987 & 1.0000 & 0.9987 \\
Masked rationales on wrongly predicted examples & 0.9780 & 0.9981 & 0.9790 \\
\bottomrule
\end{tabular}
\end{table*}

\paragraph{Predicting the final verdict from the rationale.}
We next test whether the rationale alone contains enough information to recover \textsc{Reader}'s final verdict. To avoid using \textsc{Reader}'s internal representations, we use an external BERT encoder to predict \textsc{Reader}'s final label from the rationale representation. We also evaluate a masked setting, where explicit AI/HUMAN label words in the rationale are removed before encoding. This masked setting reduces the possibility that the classifier simply reads off the final label token.

Table~\ref{tab:rationale_predictiveness} shows that the rationales are highly predictive of \textsc{Reader}'s final verdict. Even after masking explicit label words, the classifier achieves 0.9982 accuracy and 1.0000 AUROC. This indicates that the rationales contain label-supporting linguistic information beyond explicit AI/HUMAN tokens. We emphasize that this analysis measures predictiveness and alignment with the final verdict.

Overall, these diagnostics show that \textsc{Reader}'s rationales are not merely decorative output fields. They are output-consistent with the final verdict, highly predictive of the verdict even after explicit label words are masked, and associated with interpretable label-supporting cue patterns. They also exhibit weaker alignment and higher uncertainty in incorrect predictions. These findings support the view that \textsc{Reader} provides prediction-aligned, human-readable rationales alongside its classifications.

\FloatBarrier
\section{Reasoning Examples}\label{sample_analysis}
This section presents complete comparison examples between \textsc{Reader} and other flagship LLMs. These examples complement the aggregate comparison in Section~\ref{subsec:generalLLM} and Table~\ref{tab:accuracy_format_unusable_rates}. They illustrate that \textsc{Reader} often produces specific rationales that are aligned with its final prediction and grounded in linguistic cues relevant to AI-text detection, such as repetitive structure, generic phrasing, unnatural transitions, overly polished style, or lack of personal style. All models are prompted identically on the same text, using Prompt~\ref{app:prompt-balanced}. Among the 1,200 test samples, 104 are misclassified by all flagship models; \textsc{Reader} is correct on 97 of these cases, achieving $93.27\%$ accuracy on this challenging subset. Tables~\ref{tab:sample_claude_haiku_4_5_1134}, \ref{tab:sample_Gemini_1_5_Pro_1456}, and \ref{tab:sample_deepseek_v3_2_1000} show AI-generated examples from this challenging subset, while Tables~\ref{tab:sample_human_140}, \ref{tab:sample_human_1004}, and \ref{tab:sample_human_280} show human-written examples. Together, these cases span multiple AI generators and human-written sources across diverse domains, highlighting the robustness of \textsc{Reader}. 

\begin{table*}[!tbp]
\centering
\small
\renewcommand{\arraystretch}{1.2}
\caption{AI-generated text from \textit{Entertainment}}
\label{tab:sample_claude_haiku_4_5_1134}

\begin{tabular}{>{\raggedright\arraybackslash}p{2.5cm} p{12cm}}
\toprule
\rowcolor{headerblue!20}
\textbf{Source} & \textbf{Claude-Haiku-4.5} \\
\midrule
\textbf{Original Text} & I can't believe people are actually praising this movie – are they talking about something else? I went in braced for it to be bad, but honestly, the script (or whatever you want to call it) makes the whole thing feel pretty pointless. That said, the actors and crew do solid work, which is cool, but you can't polish a turd, you know? If you're a Ben G fan, literally watch any of his other movies. Dorothy S fans? Just stick with Galaxina. Peter B fans? Go watch Last Picture Show or Target instead. And if you're looking for cheap laughs at people's expense, honestly just read Peter B's hilariously terrible book, Killing of the Unicorn – it's way better. \\
\midrule
\rowcolor{headerblue!10}
\multicolumn{2}{c}{\textbf{Model Predictions and Rationales}} \\
\midrule

\rowcolor{lightgray}
\raisebox{-0.2\height}{\includegraphics[height=1em]{icons/reader.png}} \textbf{\textsc{Reader}} & \textcolor{correctgreen}{\textbf{AI \checkmark}} \newline \textit{First, the tone is hyperconfident and conversational with colloquialisms ('you know?') and abrupt transitions, patterns common in AI-generated outputs. Second, there are factual mismatches and odd recommendations (e.g., suggesting viewers 'literally watch any of his other movies' for Ben G) indicating weak real-world grounding typical of AI coherence errors. Moreover, the writing is polished yet generic, lacking personal nuance or specific anecdotes that a human reviewer would likely include. For these reasons I judge the passage to be AI} \\

\raisebox{-0.2\height}{\includegraphics[height=1em]{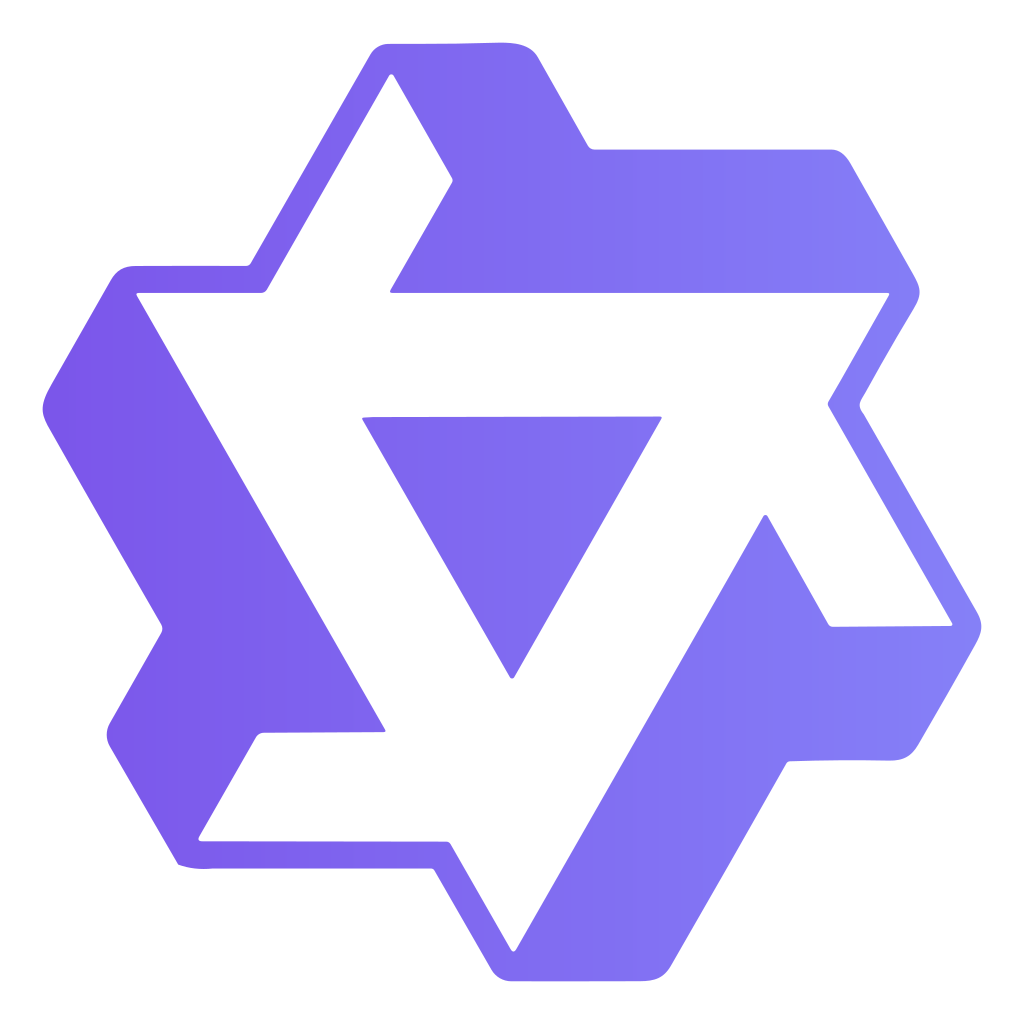}} \textbf{Qwen3-Max} & \textcolor{wrongred}{\textbf{HUMAN $\times$}} \newline \textit{First, the text has a conversational and informal tone, which is common in human writing. Second, it includes specific references to actors and movies, suggesting personal opinion and experience. Moreover, the use of colloquial phrases like 'can't polish a turd' and 'hilariously terrible book' indicates a natural, human-like expression. Finally, the text shows a clear and coherent personal opinion, which is typical of human-generated content.} \\

\raisebox{-0.2\height}{\includegraphics[height=1em]{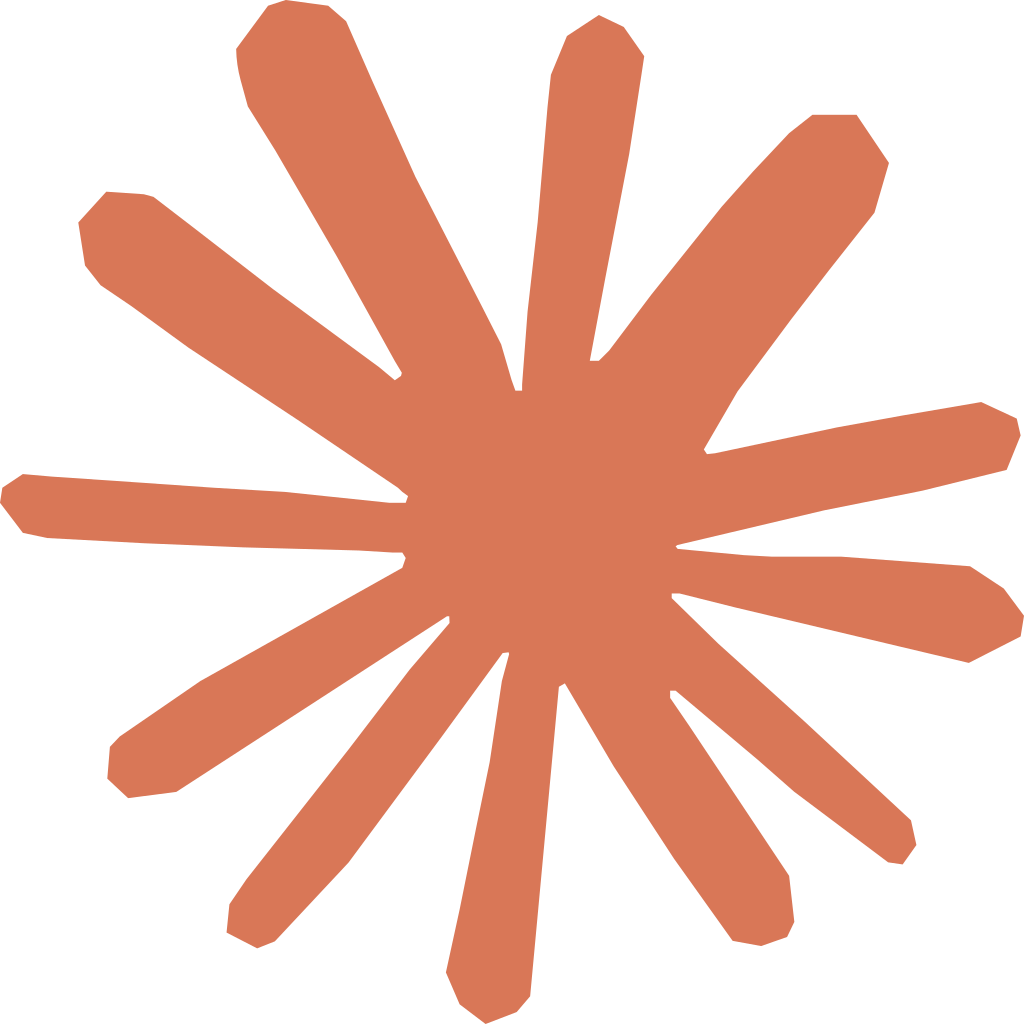}} \textbf{Claude-Sonnet-4.5} & \textcolor{wrongred}{\textbf{HUMAN $\times$}} \newline \textit{First, this text exhibits strongly human-like informal language patterns, including colloquialisms such as 'you can't polish a turd' and 'you know?' that feel naturally conversational. Second, the writer uses abbreviated celebrity names (Ben G, Dorothy S, Peter B) in a casual, insider manner typical of human film enthusiasts discussing familiar figures within their community. Moreover, the stream-of-consciousness structure with parenthetical asides ('or whatever you want to call it') and shifting focus demonstrates organic human thought patterns rather than AI's typically more organized presentation. Additionally, the specific cultural references to obscure films like 'Galaxina' and 'Killing of the Unicorn' reflect genuine personal knowledge rather than AI's tendency toward more mainstream examples. The sarcastic, slightly bitter tone throughout maintains consistent emotional authenticity that AI struggles to replicate convincingly.} \\

\raisebox{-0.2\height}{\includegraphics[height=1em]{icons/openai.png}} \textbf{GPT-5.2} & \textcolor{wrongred}{\textbf{HUMAN $\times$}} \newline \textit{First, the text has a highly opinionated, conversational voice with rhetorical questions and idioms (''you can't polish a turd''), which is more typical of spontaneous human commentary than default AI neutrality. Second, it includes specific, slightly messy cultural references and informal shorthand (''Ben G,'' ''Dorothy S,'' ''Peter B'') that read like a real reviewer assuming shared context rather than an AI trying to be maximally clear. Moreover, the structure is a natural rant with abrupt pivots and asides (''or whatever you want to call it''), not a polished, evenly balanced critique. Finally, the snarky, targeted recommendation list and the punchline about a ''hilariously terrible book'' feel like personal taste and humor rather than templated generation, so the decision is HUMAN.} \\

\raisebox{-0.2\height}{\includegraphics[height=1em]{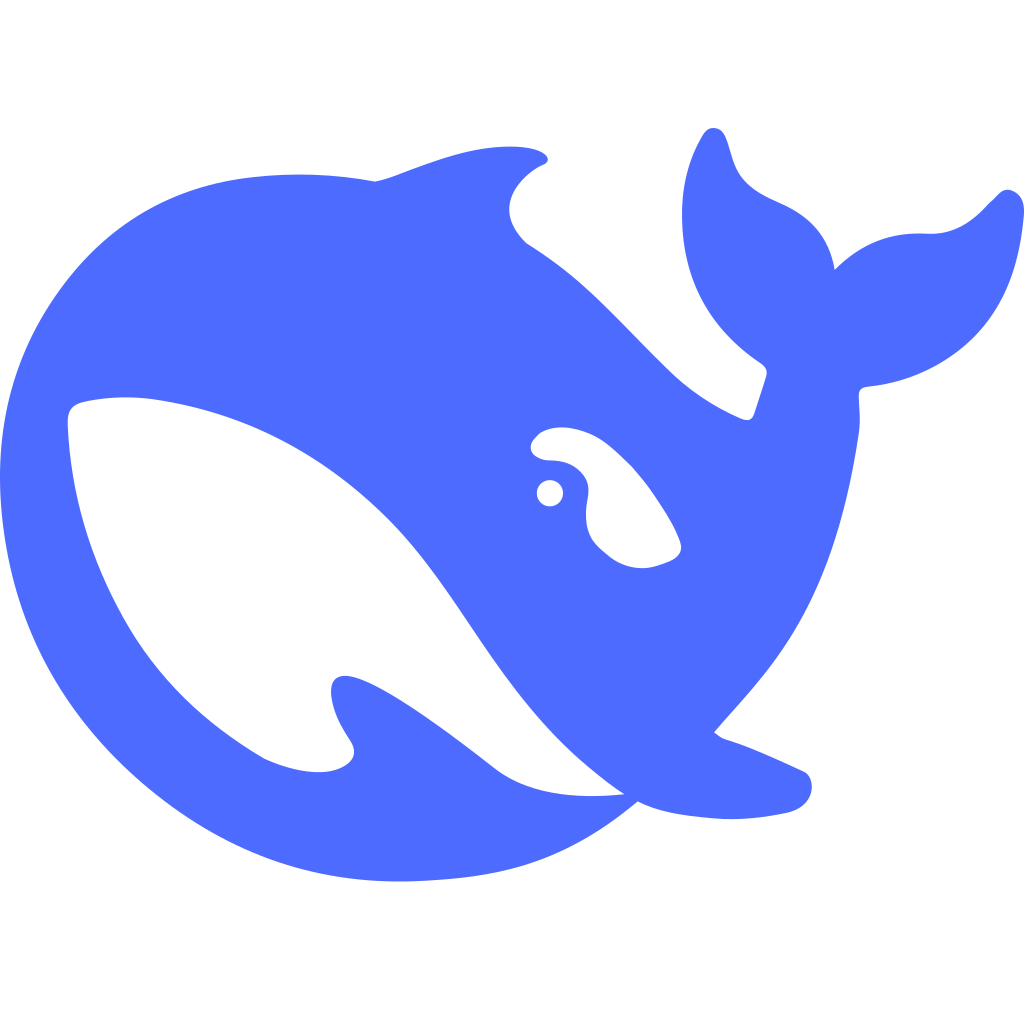}} \textbf{DeepSeek-V3.2} & \textcolor{wrongred}{\textbf{HUMAN $\times$}} \newline \textit{First, the text exhibits strong personal voice with colloquial expressions like 'you can't polish a turd' and 'literally watch any of his other movies,' which are characteristic of human conversational tone. Second, it contains specific cultural references (Galaxina, Last Picture Show, Killing of the Unicorn) and uses informal punctuation like dashes and parentheses in ways that feel organically human. Moreover, the argument builds through personal opinion ('I went in braced for it to be bad') rather than balanced analysis, and ends with a sarcastic recommendation that reflects subjective humor. Finally, the flow includes abrupt shifts and emphatic phrasing ('honestly just read') that align with human-written critique rather than AI's typically more structured output.} \\

\raisebox{-0.2\height}{\includegraphics[height=1em]{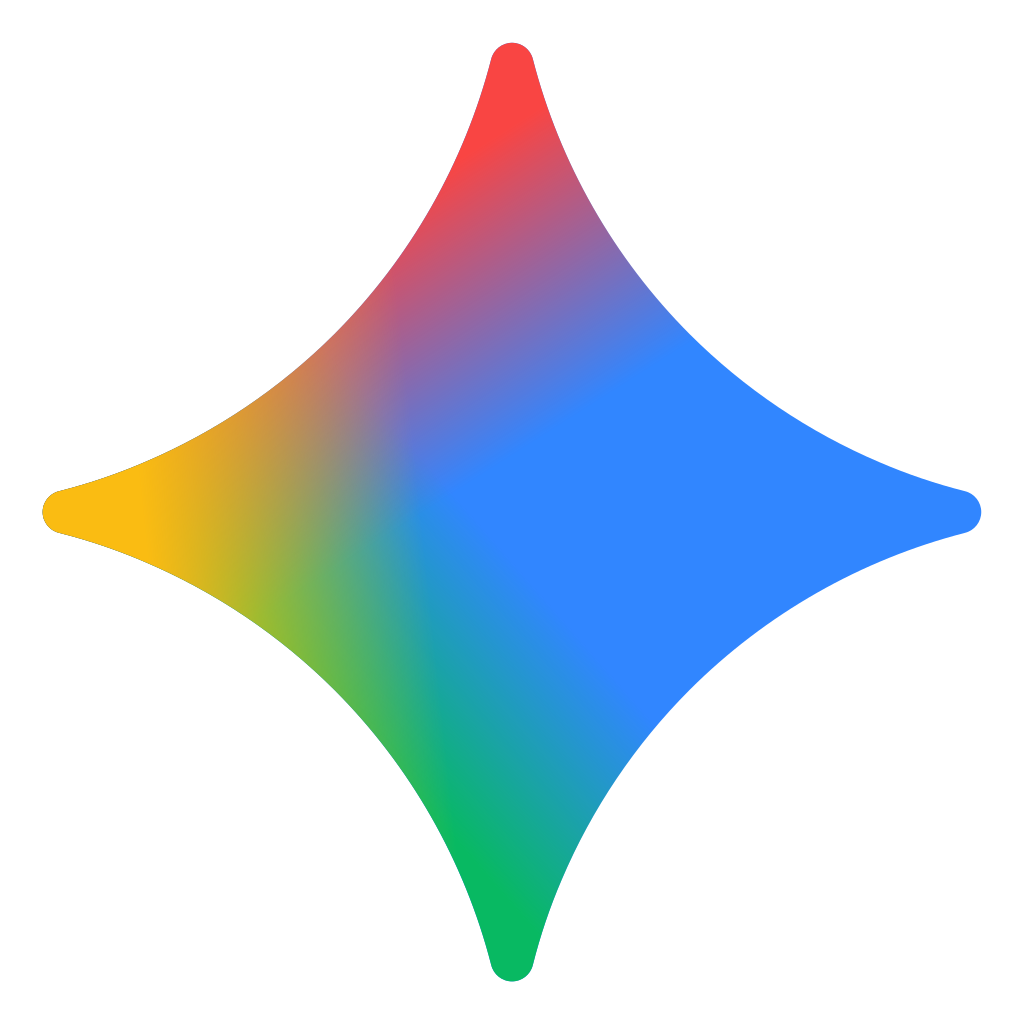}} \textbf{Gemini-3-Pro} & \textcolor{wrongred}{\textbf{HUMAN $\times$}} \newline \textit{First, the text employs a highly colloquial and informal tone, featuring idioms like 'polish a turd' and rhetorical questions that feel authentic to human speech patterns. Second, the specific references to niche film history, such as Dorothy Stratten's 'Galaxina' and Peter Bogdanovich's book, are woven into the narrative with a personal, opinionated flair. Moreover, the use of shorthand for names and expressive parentheticals suggests a specific individual's voice rather than the generic structure typical of AI. Consequently, the combination of niche knowledge and raw emotional affect confirms its human origin.} \\

\bottomrule
\end{tabular}
\end{table*}

\begin{table*}[!tbp]
\centering
\small
\renewcommand{\arraystretch}{1.2}
\caption{AI-generated text from \textit{FoodCuisine}}
\label{tab:sample_Gemini_1_5_Pro_1456}

\begin{tabular}{>{\raggedright\arraybackslash}p{2.5cm} p{12cm}}
\toprule
\rowcolor{headerblue!20}
\textbf{Source} & \textbf{Gemini-1.5-Pro} \\
\midrule
\textbf{Original Text} & This is one of the few high-quality foods my cat with fish and beef allergies can eat! It surpasses even prescription diets! Both my cats are healthy, active, and have beautiful coats. Highly recommend!

 \\
\midrule
\rowcolor{headerblue!10}
\multicolumn{2}{c}{\textbf{Model Predictions and Rationales}} \\
\midrule
\rowcolor{lightgray}
\raisebox{-0.2\height}{\includegraphics[height=1em]{icons/reader.png}} \textbf{\textsc{Reader}} & \textcolor{correctgreen}{\textbf{AI \checkmark}} \newline \textit{First, the text uses highly polished, generic positive language ('highly recommend', 'beautiful coats') typical of AI-generated promotional copy. Second, it lacks concrete specifics (no brand, ingredients, or personal details) and repeats common review patterns AI often generates. Moreover, there is a slight inconsistency—praising taste while emphasizing convenience—suggesting algorithmic tendency to balance multiple angles. Based on these signals, I judge it to be AI} \\

\raisebox{-0.2\height}{\includegraphics[height=1em]{icons/qwen.png}} \textbf{Qwen3-Max} & \textcolor{wrongred}{\textbf{HUMAN $\times$}} \newline \textit{First, the text is concise and to the point, which is typical of human-written reviews. Second, the use of exclamation marks and the personal tone (e.g., 'my cat') suggest a genuine, emotional response, which is more characteristic of human writing. Moreover, the specific mention of the cat's health and coat condition adds a level of detail that is often found in human-generated content.} \\

\raisebox{-0.2\height}{\includegraphics[height=1em]{icons/claude.png}} \textbf{Claude-Sonnet-4.5} & \textcolor{wrongred}{\textbf{HUMAN $\times$}} \newline \textit{First, the text exhibits enthusiastic, personal testimonial language with multiple exclamation marks, which is common in genuine customer reviews. Second, the specificity of mentioning 'fish and beef allergies' reflects authentic personal experience rather than generic AI-generated content. Moreover, the brief, punchy sentence structure and the use of casual phrasing like 'Highly recommend!' mirrors typical human review patterns on e-commerce platforms. The emotional investment shown through excitement markers and the comparative statement about prescription diets suggests genuine satisfaction from a real pet owner. Finally, the slightly informal yet coherent style is characteristic of authentic consumer feedback rather than the more balanced, explanatory tone typical of AI-generated reviews.} \\

\raisebox{-0.2\height}{\includegraphics[height=1em]{icons/openai.png}} \textbf{GPT-5.2} & \textcolor{wrongred}{\textbf{HUMAN $\times$}} \newline \textit{First, the text reads like a natural consumer review with specific personal context (a cat with fish and beef allergies) that grounds it in lived experience. Second, the phrasing includes casual emphasis and subjective comparisons (''surpasses even prescription diets!'') typical of human product feedback rather than a structured AI summary. Moreover, the short, enthusiastic sentences and minor redundancy (mentioning both cats separately) resemble spontaneous human writing rather than optimized AI prose. Therefore, the overall style and specificity suggest it is human-written.} \\

\raisebox{-0.2\height}{\includegraphics[height=1em]{icons/deepseek.png}} \textbf{DeepSeek-V3.2} & \textcolor{wrongred}{\textbf{HUMAN $\times$}} \newline \textit{First, the text exhibits natural conversational phrasing with exclamation points and personal enthusiasm that feels authentically human. Second, it contains specific personal details about a cat with fish and beef allergies and comparison to prescription diets, which suggests lived experience rather than generic product description. Moreover, the structure is brief yet emotionally expressive with phrases like 'highly recommend!' that mirror genuine consumer testimonials. Finally, the informal yet coherent flow without overly polished or formulaic language aligns more with spontaneous human writing than AI-generated content.} \\

\raisebox{-0.2\height}{\includegraphics[height=1em]{icons/gemini.png}} \textbf{Gemini-3-Pro} & \textcolor{wrongred}{\textbf{HUMAN $\times$}} \newline \textit{First, the text identifies a specific and realistic niche problem—fish and beef allergies—which is a common detail in authentic pet owner reviews. Second, the comparison to prescription diets suggests a personal history with the product category that goes beyond generic AI-generated descriptions. Moreover, the minor shift from focusing on one cat's allergies to the health of 'both' cats reflects a natural, human narrative flow. Finally, the concise, enthusiastic structure and standard closing recommendation are hallmark characteristics of genuine user-generated content found on e-commerce platforms.} \\

\bottomrule
\end{tabular}
\end{table*}
\vspace{-1pt}
\begin{table*}[!tbp]
\centering
\small
\renewcommand{\arraystretch}{1.2}
\caption{AI-generated text from \textit{Entertainment}}
\label{tab:sample_deepseek_v3_2_1000}

\begin{tabular}{>{\raggedright\arraybackslash}p{2.5cm} p{12cm}}
\toprule
\rowcolor{headerblue!20}
\textbf{Source} & \textbf{DeepSeek-V3.2} \\
\midrule
\textbf{Original Text} & It’s strange, but all of this reminds me: he never really had the permission to do that. He didn’t earn it. He’s had plenty of time, but no—other things always mattered more. Honestly, it would take him twice as long as he’s even known me to deserve that right.
I know this probably doesn’t make sense to anyone else. Maybe that’s for the best.\\
\midrule
\rowcolor{headerblue!10}
\multicolumn{2}{c}{\textbf{Model Predictions and Rationales}} \\
\midrule
\rowcolor{lightgray}
\raisebox{-0.2\height}{\includegraphics[height=1em]{icons/reader.png}} \textbf{\textsc{Reader}} & \textcolor{correctgreen}{\textbf{AI \checkmark}} \newline \textit{First, the passage uses generic, evenly balanced phrasing and lacks concrete sensory detail or a distinct personal voice, which often indicates AI generation. Second, there are mechanical patterns (repetition of 'even known me' and hedging like 'Maybe that's for the best') common in AI outputs. Moreover, transitions are overly smooth and neutral rather than showing human-like idiosyncratic errors or emotional specificity. Based on these signals, I judge the text to be AI} \\

\raisebox{-0.2\height}{\includegraphics[height=1em]{icons/qwen.png}} \textbf{Qwen3-Max} & \textcolor{wrongred}{\textbf{HUMAN $\times$}} \newline \textit{First, the text has a stream-of-consciousness style with a personal and introspective tone, which is more characteristic of human writing. Second, the sentences are somewhat fragmented and convey a sense of internal monologue, suggesting a natural, human-like thought process. Moreover, the content is emotionally charged and contextually vague, which is typical of human expression rather than AI-generated text.} \\

\raisebox{-0.2\height}{\includegraphics[height=1em]{icons/claude.png}} \textbf{Claude-Sonnet-4.5} & \textcolor{wrongred}{\textbf{HUMAN $\times$}} \newline \textit{First, this text exhibits highly personal, fragmented thinking with deliberately vague references ('that,' 'it') that assumes insider knowledge, which is characteristic of authentic human journal-style writing. Second, the emotional undertone is subtly resentful and passive-aggressive ('he never really had the permission,' 'He didn't earn it'), expressed through natural, conversational syntax rather than polished prose. Moreover, the self-aware admission 'I know this probably doesn't make sense to anyone else' demonstrates meta-cognition about the opacity of the writing\u2014a very human acknowledgment of communicating something intentionally cryptic. The informal tone ('Honestly,' 'no\u2014other things'), the use of an em dash for interrupted thought, and the deliberate withholding of context all point to spontaneous human expression rather than AI's typical tendency to provide clear, complete explanations.} \\

\raisebox{-0.2\height}{\includegraphics[height=1em]{icons/openai.png}} \textbf{GPT-5.2} & \textcolor{wrongred}{\textbf{HUMAN $\times$}} \newline \textit{First, the passage reads like introspective, emotionally charged interior monologue with idiosyncratic judgments ("he never really had the permission," "didn't earn it"), which is common in human diary-like writing. Second, the syntax shows natural irregularities---fragmented pauses, em-dash interruption ("but no---"), and a slightly awkward comparative construction ("twice as long as he's even known me") that feels unpolished rather than optimally smoothed. Moreover, the text is compact and context-dependent, withholding specifics in a way that suggests a real speaker protecting private details rather than an AI optimizing clarity. Therefore, the features overall more strongly indicate a human author.} \\

\raisebox{-0.2\height}{\includegraphics[height=1em]{icons/deepseek.png}} \textbf{DeepSeek-V3.2} & \textcolor{wrongred}{\textbf{HUMAN $\times$}} \newline \textit{First, the text displays emotional complexity and personal reflection that feels authentically human, with phrases like 'It's strange, but all of this reminds me' suggesting genuine introspection. Second, the narrative includes subtle contradictions and unresolved thoughts ('I know this probably doesn\u2019t make sense to anyone else'), which are characteristic of human stream-of-consciousness rather than AI's typically more structured output. Moreover, the conversational tone and use of colloquial expressions ('Honestly,' 'Maybe that\u2019s for the best') create an intimate, subjective voice that AI often struggles to replicate convincingly. Finally, the text lacks the overly polished, formulaic phrasing common in AI-generated content, instead embracing a raw, fragmented quality that mirrors human thought processes.} \\

\raisebox{-0.2\height}{\includegraphics[height=1em]{icons/gemini.png}} \textbf{Gemini-3-Pro} & \textcolor{wrongred}{\textbf{HUMAN $\times$}} \newline \textit{First, the text utilizes a deeply personal and introspective voice that relies on contextual ambiguity, a characteristic often found in human-authored creative writing or personal journals. Second, the rhythmic use of the em-dash and short, punchy sentences suggests a naturalistic internal monologue rather than the more balanced or explanatory structures typically produced by AI. Moreover, the specific logical phrasing regarding time and relationship duration adds a layer of emotional complexity that feels distinctly grounded in human experience. Therefore, the lack of generic fillers or a structured conclusion further points to a human origin.} \\
\bottomrule
\end{tabular}
\end{table*}

\begin{table*}[!tbp]
\centering
\small
\renewcommand{\arraystretch}{1.2}
\caption{Human-written text from \textit{ROC}}
\label{tab:sample_human_140}

\begin{tabular}{>{\raggedright\arraybackslash}p{2.5cm} p{12cm}}
\toprule
\rowcolor{headerblue!20}
\textbf{Source} & \textbf{Human} \\
\midrule
\textbf{Original Text} & Kate listened to loud music on her laptop. Her roommate asked her to turn it down. Instead, Kate decided to wear headphones. The music was still too annoying for the roommate. Kate told the roommate to just wear earplugs and deal with it. \\
\midrule
\rowcolor{headerblue!10}
\multicolumn{2}{c}{\textbf{Model Predictions and Rationales}} \\
\midrule
\rowcolor{lightgray}
\raisebox{-0.2\height}{\includegraphics[height=1em]{icons/reader.png}} \textbf{\textsc{Reader}} & \textcolor{correctgreen}{\textbf{HUMAN \checkmark}} \newline \textit{First, the passage uses very short, colloquial sentences and a personal conversational tone (e.g., 'just wear earplugs and deal with it'), which is common in human casual writing. Second, there are small stylistic variations and a lack of excessive smoothing or explanatory padding often seen in AI outputs, suggesting human spontaneity. Moreover, the narrative is focused and anecdotal without mechanical transitions, reinforcing a human origin. Therefore I judge this text as HUMAN.} \\

\raisebox{-0.2\height}{\includegraphics[height=1em]{icons/qwen.png}} \textbf{Qwen3-Max} & \textcolor{wrongred}{\textbf{AI $\times$}} \newline \textit{First, the text is straightforward and lacks complex sentence structures, which is common in both AI and human writing. Second, the narrative flow is logical and coherent, but the resolution of the conflict (Kate's suggestion to wear earplugs) feels abrupt and somewhat unsympathetic, which can be a characteristic of AI-generated text. Moreover, the lack of nuanced dialogue or deeper character development points towards an AI-generated origin.} \\

\raisebox{-0.2\height}{\includegraphics[height=1em]{icons/claude.png}} \textbf{Claude-Sonnet-4.5} & \textcolor{wrongred}{\textbf{AI $\times$}} \newline \textit{First, the text exhibits a simple, linear narrative structure with straightforward cause-and-effect relationships that could indicate either origin. Second, the dialogue and conflict resolution are unusually illogical and lack natural human reasoning—Kate switching to headphones should solve the noise problem, yet the roommate still complains about music being 'too annoying,' which makes no physical sense. Moreover, the abrupt, dismissive ending where Kate tells her roommate to wear earplugs contradicts the earlier attempt at compromise, suggesting a formulaic story structure rather than authentic human interaction. The vocabulary is basic and the sentences follow a repetitive subject-verb-object pattern typical of AI-generated short narratives. Therefore, this appears to be AI-generated text.} \\

\raisebox{-0.2\height}{\includegraphics[height=1em]{icons/openai.png}} \textbf{GPT-5.2} & \textcolor{wrongred}{\textbf{AI $\times$}} \newline \textit{First, the text uses very short, uniformly structured sentences with simple subject-verb-object patterns, which is common in AI-generated or instructional prose. Second, the narrative is generic and lacks distinctive sensory detail, dialogue, or idiosyncratic phrasing that often appears in natural human storytelling. Moreover, the progression reads like a schematic moral/behavior vignette (problem → request → partial fix → conflict) with minimal stylistic variation. Therefore, these features collectively suggest the passage is AI-generated.} \\

\raisebox{-0.2\height}{\includegraphics[height=1em]{icons/deepseek.png}} \textbf{DeepSeek-V3.2} & \textcolor{wrongred}{\textbf{AI $\times$}} \newline \textit{First, the text exhibits an overly simplistic and formulaic sentence structure with repetitive subject-verb-object patterns. Second, the narrative lacks natural human conversational details, emotional nuance, or descriptive elements that would typically accompany such an interpersonal conflict. Moreover, the resolution is presented in an unrealistically abrupt and mechanical manner without the hesitation, justification, or emotional tone a human writer would likely include. Finally, the entire passage reads like a sterile sequence of events constructed to demonstrate cause-and-effect rather than to convey a genuine human experience.} \\

\raisebox{-0.2\height}{\includegraphics[height=1em]{icons/gemini.png}} \textbf{Gemini-3-Pro} & \textcolor{wrongred}{\textbf{AI $\times$}} \newline \textit{First, the text employs a repetitive and stilted sentence structure, where every sentence follows a basic subject-verb-object pattern typical of simple AI generation. Second, the lack of cohesive transitions between sentences creates a disjointed rhythm often seen in machine-generated summaries. Moreover, there is a minor logical inconsistency regarding why the music remains 'annoying' after headphones are used, suggesting a formulaic approach to conflict rather than human-like narrative nuance. Finally, the repetitive use of nouns like 'Kate' and 'roommate' instead of varied pronouns is a common hallmark of AI-generated text.} \\

\bottomrule
\end{tabular}
\end{table*}

\begin{table*}[!tbp]
\centering
\small
\renewcommand{\arraystretch}{1.2}
\caption{Human-written text from \textit{TLDR}}
\label{tab:sample_human_1004}

\begin{tabular}{>{\raggedright\arraybackslash}p{2.5cm} p{12cm}}
\toprule
\rowcolor{headerblue!20}
\textbf{Source} & \textbf{Human} \\
\midrule
\textbf{Original Text} & Apple, Google, and Amazon have formed a group called Project Connected Home over IP. The group aims to develop an open-source smart home standard to ensure that any supported smart home device will work with any voice assistant. For developers, this means that product development will be simplified and costs will be reduced. The new standard will work alongside existing connectivity protocols. While not all devices will connect directly to the internet, the new standard will be based on Internet Protocol as it is ideal for sending messages between devices securely and privately. The group will initially focus on physical safety devices before moving onto other consumer and commercial devices. \\
\midrule
\rowcolor{headerblue!10}
\multicolumn{2}{c}{\textbf{Model Predictions and Rationales}} \\
\midrule
\rowcolor{lightgray}
\raisebox{-0.2\height}{\includegraphics[height=1em]{icons/reader.png}} \textbf{\textsc{Reader}} & \textcolor{correctgreen}{\textbf{HUMAN \checkmark}} \newline \textit{First, the passage has a concise, journalistic tone with specific details (Project Connected Home over IP, initial focus on physical safety devices) suggesting human summarization. Second, it uses varied sentence lengths, includes a personal 'This means...' clause, and lacks common AI markers like excessive hedging or mechanical repetition. Moreover, it presents a coherent, plausible viewpoint with contextual grounding rather than generic filler, which is typical of human writing. Therefore I judge this text to be HUMAN.} \\

\raisebox{-0.2\height}{\includegraphics[height=1em]{icons/qwen.png}} \textbf{Qwen3-Max} & \textcolor{wrongred}{\textbf{AI $\times$}} \newline \textit{First, the text is well-structured and logically flows, which is characteristic of both AI and human writing. Second, the content is informative and concise, typical of a professional or technical summary, which could be written by either. Moreover, the language used is straightforward and free of colloquialisms, which is more indicative of AI-generated text. Finally, the text does not contain any unique or creative phrasing that would strongly suggest a human touch.} \\

\raisebox{-0.2\height}{\includegraphics[height=1em]{icons/claude.png}} \textbf{Claude-Sonnet-4.5} & \textcolor{wrongred}{\textbf{AI $\times$}} \newline \textit{First, the text exhibits a highly structured, encyclopedic tone that prioritizes information delivery over stylistic variation, which is common in AI-generated content. Second, the sentence construction follows a predictable pattern with consistent subject-verb-object formations and lacks the natural variations typical of human writing. Moreover, the text presents information in a perfectly logical sequence without tangential observations or personal commentary that human writers often include. Additionally, the language is uniformly formal and technical without any colloquialisms, rhetorical questions, or engaging transitions that would suggest human authorship. Finally, the absence of any redundancy, self-correction, or emphatic phrasing further indicates AI generation. Based on these characteristics, this text appears to be AI-generated.} \\

\raisebox{-0.2\height}{\includegraphics[height=1em]{icons/openai.png}} \textbf{GPT-5.2} & \textcolor{wrongred}{\textbf{AI $\times$}} \newline \textit{First, the passage is highly neutral and informational, with polished, newsroom-style phrasing and no personal perspective or idiosyncratic voice. Second, it follows a very linear, checklist-like structure (who/what/goal/benefits/how/limitations/roadmap) that reads like an autogenerated summary. Moreover, it relies on broad, generic claims (''simplified,'' ''costs will be reduced,'' ''securely and privately'') without concrete examples, quotes, or specific sourcing that human writers often include. Therefore, the overall tone and uniform sentence construction suggest AI-generated text.} \\

\raisebox{-0.2\height}{\includegraphics[height=1em]{icons/deepseek.png}} \textbf{DeepSeek-V3.2} & \textcolor{wrongred}{\textbf{AI $\times$}} \newline \textit{First, the text exhibits a highly structured and formulaic organization typical of AI-generated news summaries, moving systematically from announcement to purpose to technical details. Second, the language is consistently neutral and informational without any personal voice, stylistic flourishes, or subjective commentary that would suggest human authorship. Moreover, the content presents complex technical concepts (like Internet Protocol and connectivity protocols) with perfect clarity and concision but no depth of analysis or unique perspective. Finally, the paragraph flows with a predictable cause-and-effect logic and lacks the subtle digressions, hedging, or conversational markers often found in human writing on technical topics.} \\

\raisebox{-0.2\height}{\includegraphics[height=1em]{icons/gemini.png}} \textbf{Gemini-3-Pro} & \textcolor{wrongred}{\textbf{AI $\times$}} \newline \textit{First, the text maintains a consistently neutral and objective tone that lacks the individualistic style or varied prosody often found in human writing. Second, the structural organization is highly formulaic, progressing logically from the announcement to the benefits and then to the technical specifications. Moreover, the phrasing 'this means that' and the precise technical summaries suggest a synthetic synthesis of information typical of language models. Finally, the absence of any subjective nuance or stylistic 'noise' points toward an AI-generated summary of technical news.} \\
\bottomrule
\end{tabular}
\end{table*}

\begin{table*}[!tbp]
\centering
\small
\renewcommand{\arraystretch}{1.2}
\caption{Human-written text from \textit{SQuAD}}
\label{tab:sample_human_280}

\begin{tabular}{>{\raggedright\arraybackslash}p{2.5cm} p{12cm}}
\toprule
\rowcolor{headerblue!20}
\textbf{Source} & \textbf{Human} \\
\midrule
\textbf{Original Text} & Oklahoma is the 20th largest state in the United States, covering an area of 69,898 square miles (181,035 km2), with 68,667 square miles (177847 km2) of land and 1,281 square miles (3,188 km2) of water. It is one of six states on the Frontier Strip and lies partly in the Great Plains near the geographical center of the 48 contiguous states. It is bounded on the east by Arkansas and Missouri, on the north by Kansas, on the northwest by Colorado, on the far west by New Mexico, and on the south and near-west by Texas. \\
\midrule
\rowcolor{headerblue!10}
\multicolumn{2}{c}{\textbf{Model Predictions and Rationales}} \\
\midrule
\rowcolor{lightgray}
\raisebox{-0.2\height}{\includegraphics[height=1em]{icons/reader.png}} \textbf{\textsc{Reader}} & \textcolor{correctgreen}{\textbf{HUMAN \checkmark}} \newline \textit{First, the passage has a neutral, encyclopedia-like tone with precise numeric data and parenthetical metric conversions, suggesting human-edited source material. Second, its specific reference to the Frontier Strip and exact bordering states matches known human-written reference text rather than generic AI filler. Moreover, there are no clear AI telltales like excessive hedging, unnatural repetition, or mechanical transitions; instead it shows small stylistic variation typical of human prose. Based on these points, I judge it to be HUMAN.} \\

\raisebox{-0.2\height}{\includegraphics[height=1em]{icons/qwen.png}} \textbf{Qwen3-Max} & \textcolor{wrongred}{\textbf{AI $\times$}} \newline \textit{First, the text is highly factual and structured, which is a common characteristic of AI-generated content. Second, the precision in the use of numbers and geographical details suggests a level of accuracy that is often associated with AI. Moreover, the sentence structure and flow are consistent and lack the personal touch or slight variations that human writing might include.} \\

\raisebox{-0.2\height}{\includegraphics[height=1em]{icons/claude.png}} \textbf{Claude-Sonnet-4.5} & \textcolor{wrongred}{\textbf{AI $\times$}} \newline \textit{First, the text exhibits a highly encyclopedic and formulaic structure, presenting geographic facts in a systematic manner typical of reference material. Second, the precise numerical data (including dual unit measurements in square miles and kilometers with exact conversions) and comprehensive border state listings follow a template-like pattern characteristic of Wikipedia or similar factual databases. Moreover, the writing style is impersonal, objective, and devoid of any subjective commentary or stylistic flourishes that would suggest individual human authorship. The mechanical enumeration of boundaries using directional qualifiers ('on the east,' 'on the north') further reinforces a standardized, reference-based composition. While humans certainly write encyclopedia entries, this particular passage's sterile precision and structural rigidity suggest it was either directly copied from or generated to mimic reference material, most likely AI-generated to reproduce encyclopedic content.} \\

\raisebox{-0.2\height}{\includegraphics[height=1em]{icons/openai.png}} \textbf{GPT-5.2} & \textcolor{wrongred}{\textbf{AI $\times$}} \newline \textit{First, the passage is densely factual and metric-heavy, using parenthetical unit conversions and exact figures in a way typical of encyclopedic or machine-generated summaries. Second, the sentence structure is uniform and list-like, especially in the boundary description, with little narrative voice or variation. Moreover, it includes an odd-sounding phrase (“one of six states on the Frontier Strip”) that reads like a stitched fact without context, which is common in AI or automated paraphrase. Therefore, the overall tone and construction suggest AI-generated text.} \\

\raisebox{-0.2\height}{\includegraphics[height=1em]{icons/deepseek.png}} \textbf{DeepSeek-V3.2} & \textcolor{wrongred}{\textbf{AI $\times$}} \newline \textit{First, the text presents factual geographic and statistical data with precise numerical values and systematic organization. Second, it follows a structured, encyclopedic format typical of informational sources rather than personal narrative. Moreover, the language is neutral, objective, and lacks any subjective expression, personal voice, or stylistic flourishes. Finally, the content is densely packed with technical details presented in a logical sequence, resembling automated compilation rather than human storytelling.} \\

\raisebox{-0.2\height}{\includegraphics[height=1em]{icons/gemini.png}} \textbf{Gemini-3-Pro} & \textcolor{wrongred}{\textbf{AI $\times$}} \newline \textit{First, the text adopts a formulaic and dry encyclopedic style that is characteristic of AI-generated factual summaries. Second, the consistent use of parenthetical conversions for every unit of measurement indicates a systematic approach to data presentation often seen in large language models. Moreover, the text lacks any subjective phrasing or stylistic idiosyncrasies, following a predictable structural template from general dimensions to specific cardinal boundaries. Finally, the presence of minor formatting inconsistencies, such as a missing comma in '177847' compared to other numbers, is a common artifact in AI-generated data strings.} \\

\bottomrule
\end{tabular}
\end{table*}
\FloatBarrier

\end{document}